\documentclass{article}

\usepackage[square,numbers,compress]{natbib}
\PassOptionsToPackage{numbers, compress}{natbib}

\usepackage[preprint]{neurips_2025}

\usepackage[utf8]{inputenc} 
\usepackage[T1]{fontenc}    
\usepackage{hyperref}       
\usepackage{url}            
\usepackage{booktabs}       
\usepackage{amsfonts}       
\usepackage{nicefrac}       
\usepackage{microtype}      
\usepackage[table]{xcolor}

\definecolor{reda}{RGB}{192,0,0}
 \hypersetup{
 	colorlinks   = true,
 	urlcolor     = blue,
 	linkcolor    = reda,
 	citecolor   = blue
 }

\usepackage{seqsplit}
\usepackage{fontawesome5}

\title{Multi-objective Large Language Model Alignment \\ with Hierarchical Experts}

\author{
Zhuo Li$^{1*}$,
Guodong Du$^{1*}$,
Weiyang Guo$^{1}$,
Yigeng Zhou$^{1}$,
Xiucheng Li$^{1}$, \\
\textbf{Wenya Wang}$^{2}$,
\textbf{Fangming Liu}$^{3}$,
\textbf{Yequan Wang}$^{4}$,
\textbf{Deheng Ye}$^{5}$,
\textbf{Min Zhang}$^{1}$,
\textbf{Jing Li}$^1$\textsuperscript{\faEnvelope} \\
$^{1}$Harbin Institute of Technology, Shenzhen, China \\
$^{2}$Nanyang Technological University, Singapore  \quad
$^{3}$Peng Cheng Laboratory, China \\
$^{4}$Beijing Academy of Artificial Intelligence, China  \quad
$^{5}$Tencent, China \\
\texttt{zuoer190191@mail.ustc.edu.cn} \quad \texttt{jingli.phd@hotmail.com} 
}

\usepackage[textsize=tiny]{todonotes}

\usepackage{wrapfig}
\usepackage{amsmath}
\usepackage{amssymb}
\usepackage{mathtools}
\usepackage{amsthm}

\usepackage[table]{xcolor}

\usepackage{algorithm}
\usepackage{algorithmic}

\usepackage{amsmath}
\newtheorem{theorem}{Theorem}[section]

\newtheorem{lemma}[theorem]{Lemma}

\usepackage{multirow}
\usepackage{pifont}
\usepackage{graphicx} 
\usepackage{xspace}
\newcommand{\ourapproach}{\texttt{HoE}\xspace}

\newcommand{\pub}[1]{{\color{gray}{\tiny{[{#1}]\!}}}}

\definecolor{mygray}{gray}{.9}
\definecolor{ggray}{RGB}{127,127,127}
\definecolor{reda}{RGB}{192,0,0}
\definecolor{redb}{RGB}{217,148,143}
\definecolor{myyellow}{RGB}{190,144,0}
\definecolor{mygreen}{RGB}{80,100,40}
\definecolor{myblue}{RGB}{30,90,100}
\definecolor{tabhighlight}{HTML}{e5e5e5}
\definecolor{cyan}{HTML}{00B075}

\usepackage{enumitem}

\newcommand{\ie}{\emph{i.e.,}\xspace}
\newcommand{\eg}{\emph{e.g.,}\xspace}

\begin{document}

\maketitle

\begin{abstract}
 Aligning large language models (LLMs) to simultaneously satisfy multiple objectives remains a significant challenge, especially given the diverse and often conflicting nature of human preferences.
 Existing alignment methods struggle to balance trade-offs effectively, often requiring costly retraining or yielding suboptimal results across the Pareto frontier of preferences.
 In this paper, we introduce \ourapproach (Hierarchical Mixture-of-Experts), a \textit{lightweight}, \textit{parameter-efficient}, and \textit{plug-and-play} approach that eliminates the need for model training, while enabling LLMs to adapt across the entire Pareto frontier and accommodate diverse user preferences.  
 In particular, \ourapproach consists of three hierarchical components: LoRA Experts, Router Experts and Preference Routing, reaching optimal Pareto frontiers and achieving a trade-off between parameter size, training cost, and performance. 
 We evaluate \ourapproach across various tasks on 14 objectives and 200 different preferences among 6 benchmarks, demonstrating superior performance over 15 recent baselines.
 Code is available in the supplementary materials.
\let\thefootnote\relax\footnotetext{\faEnvelope~Corresponding author. *~Equal contribution.}
\end{abstract}

\section{Introduction}
\label{introduction}

Recent advancements in large language models (LLMs)  have showcased impressive performance across a wide array of tasks~\cite{llm0, llm_training0, human_need,F2STrans_icml25,zesheng_acl25,longhui_acl25,yigeng_acl25,yifan_acl25,weiyang_acl25,kedkg_aaai_25,xianglicoling25,DBLP:conf/emnlp/MaLZ23a,junlin24acl}.
However, aligning these models to simultaneously satisfy  multiple human objectives - such as helpfulness, harmlessness, and faithfulness - remains a significant challenge~\cite{human_need,human_preference0,Linear_MORL}. 
\textit{First}, existing methods~\cite{Linear_MORL,human_preference0} are inherently limited in capturing the multi-dimensional and often conflicting nature of human values.
This limitation becomes particularly evident when attempting to satisfy a wide range of user preferences or adapt to complex scenarios where trade-offs between objectives are unavoidable~\cite{RS, RiC,steering,MODPO,PersonalizedSoups,HaM}. 
\textit{Second}, current approaches ~\cite{MORLHF,MODPO} that attempt to address this issue by training multiple models suffer from significant inefficiency in both parameter scale and training cost, making them impractical at scale.

\begin{wrapfigure}{r}{0.56\textwidth}
\begin{center}
\vspace{-5pt}
\centerline{\includegraphics[width=.56\columnwidth, trim=20 385 180 60, clip]{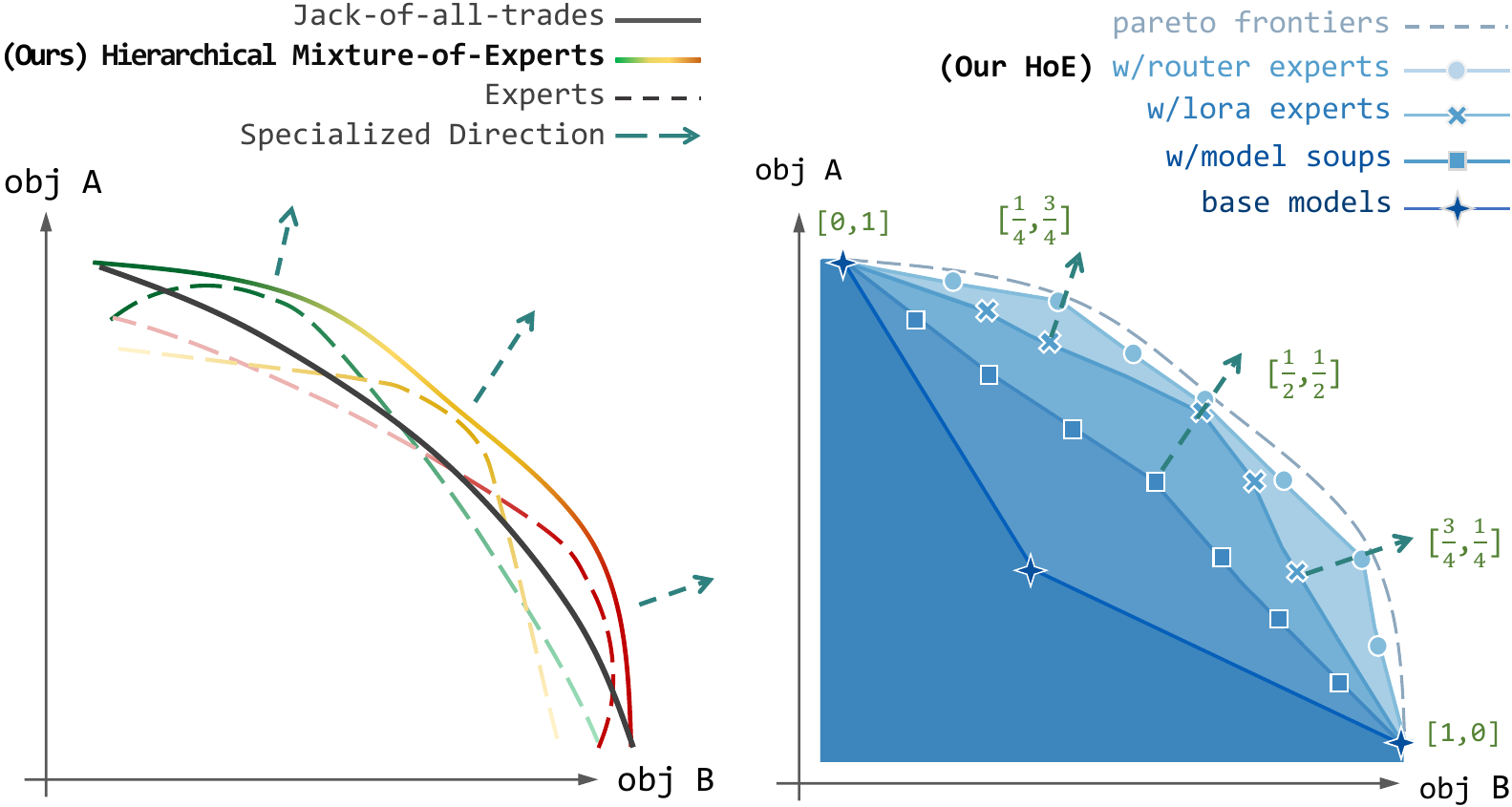}}
\vspace{5pt}
\caption{(Left) \ourapproach decomposes the multi-objective alignment problem into a series of single-preference subproblems, each handled by a specialized expert. (Right) \ourapproach employs hierarchical experts, integrating LoRA and router experts to approach the optimal Pareto frontier.}
\label{fig:motivation0}
\end{center}
\vspace{-15pt}
\end{wrapfigure}

Multi-objective alignment (MOA) requires a flexible framework capable of dynamically adapting to diverse user preferences, essentially acting as a ``jack-of-all-trades''.
However, this goal is intrinsically difficult \cite{RiC,PersonalizedSoups} due to the inherently steerable nature of multi-objective alignment (MOA).  
\textit{First}, the parameters fine-tuned for individual objectives often conflict with each other~\cite{TaskArithmeic}, reflecting the ``no free lunch'' principle~\cite{FREE-Merging} observed in multi-task learning (MTL). 
Rescaling the preference vector to favor one objective may boost that dimension but typically degrades performance on others~\cite{PCBmerging,Ties-merging,AdaMerging}. 
\textit{Second}, competition also even exists across preferences along the Pareto frontier~\cite{CLP}.
A single model can excel at only a limited set of preferences and cannot achieve optimal performance across the entire preference space~\cite{MOD,MODPO,MORLHF}. 
For instance, a model trained uniformly across all weightings (black solid Pareto frontier in Fig.~\ref{fig:motivation0}) often underperforms compared to the expert model fine-tuned preference-specifically (colored dashed Pareto frontier) at that individual preference(\eg [0.5, 0.5]) .

To overcome this limitation, we adopt a decomposition-based strategy for multi-objective alignment, breaking down the multi-objective alignment problem into a series of single-preference subproblems~\cite{MOEAD}. 
Each subproblem is handled by a specialized parameters, referred to as experts. 
These experts are each assigned to a distinct preference, focus solely on their corresponding preferences and optimize within their localized subproblem regions.  
This strategy avoids the pitfalls of a single monolithic model attempting to cover the entire Pareto frontier, thereby circumventing the steerability bottleneck. 
By constructing the full Pareto frontier from a collection of localized, preference-specific experts, our method enables precise and simplified optimization at each point along the pareto frontier.



\setlength{\abovecaptionskip}{0.cm}

One promising framework for such decomposition strategy lies in the LoRA-based Mixture-of-Experts (LoRAMoE)~\cite{LoraMoE, LoraMoE1, LoraMoE2,LoraMoE3} framework, originally developed for multi-task learning (MTL). 
LoRAMoE is a parameter-efficient approach that shares and freezes the parameters of base model and trains multiple lightweight LoRA adapters as task-specific experts. 
During inference, a router network selects and activates a subset of these experts, adapting the model's behavior dynamically. 
Despite its efficiency and modularity, the potential of LoRAMoE for multi-objective alignment remains largely untapped in current research~\cite{fastMoE} due to its unsteerable nature.
It is designed for uniform-task balancing, acts as a fixed single-preference model (\eg [0.5, 0.5] in 2-obj) when applied directly to MOA, which limits its applicability to arbitrary user preferences.
In this study, we aim to bridge this gap by fully leveraging the strengths of the Mixture-of-Experts framework and extending the LoRAMoE approach for multi-objective alignment.

In this work, we propose \ourapproach, a novel hierarchical Mixture-of-Experts framework for multi-objective alignment.
\ourapproach is a \textbf{\textit{lightweight}}, \textbf{\textit{parameter-efficient}}, and \textbf{\textit{plug-and-play}} solution that eliminates the need for training any models while achieves strong performance across the entire Pareto frontier. It combines the decomposition principle with the LoRA-based MoE design to enable scalable, efficient and fine-grained control over the entire Pareto frontier.
Specificaly, \ourapproach comprises three hierarchical components: LoRA experts, router experts, and preference routing. 
(1) ``LoRA experts'' are first extracted \textit{without training} from off-the-shelf single-objective models using task-vector singular value decomposition (task-SVD), capturing distinct alignment objectives in compact adapter modules.
(2) ``multi-objective LoRA experts'' are then synthesized, also\textit{ without training} by merging multiple existing single-objective LoRA experts, to enable on-demand generation of alignment capabilities across arbitrary preference configurations.
(3) ``Router experts'' are trained with negligible parameters to dynamically select and combine the appropriate experts based on user-specified preferences, allowing efficient traversal of the Pareto frontier.
Through this hierarchical design, \ourapproach not only provides precise control over preference-specific behavior but also balances alignment performance, parameter cost, and training efficiency. It serves as a practical and effective solution for scalable multi-objective alignment in LLMs.

\begin{table*}[ht]
\vspace{-7pt}
\centering
\renewcommand{\arraystretch}{1.27}
\caption{Comparison with other alignment methods. M is number of preference and N is number of objectives. Note that $M \gg N$. Our \ourapproach approach is a pareto-steerable and lightweight method with highest scalability, least storage cost and least inference cost, which  eliminates the need for retraining any new models or any structed prompts. Each characteristic is empirically conformed in Section~\ref{txt: Comparison with others}. }
\vspace{3pt}
\small
\resizebox{1.0\linewidth}{!}{  
\begin{tabular}{l|c|c|c|c|c|c|c}
\bottomrule
\rowcolor{mygray}
Characteristic ($\rightarrow$) & \small\textbf{Number of} & \small\textbf{Inference}& \small\textbf{Number of} & \small\textbf{Pareto} & \small \textbf{Multi-task} & &\small\textbf{Free from} \\ 
\rowcolor{mygray}
Method ($\downarrow$) & \small \textbf{stored models} & \small \textbf{cost}& \small \textbf{trained models} &\small \textbf{steerable} & \small\textbf{ability}  &\multirow{-2}{*}{ \textbf{Scalability}}  &\small\textbf{prompting}\\

\hline
\hline 
MORLHF \pub{IEEE 21} \cite{MORLHF}      &M  &1  &M  &\checkmark &\ding{55} &Retrain &\checkmark\\
MODPO \pub{ACL 24} \cite{MODPO}       &M  &1  &M  &\checkmark &\ding{55} &Retrain &\checkmark\\
RS \pub{NeurIPS 23} \cite{RS}          &N  &1  &0  &\checkmark &\checkmark &\checkmark &\checkmark\\
RiC \pub{ICML 24} \cite{RiC}          &\scalebox{0.9}{$> \!1$} &1 &\scalebox{0.9}{$> \!1$}&\checkmark &\ding{55} &Retrain &\ding{55}\\
DPA \pub{ACL 24} \cite{DPA}              &1  &1  &1  &\checkmark &\ding{55} &Retrain &\ding{55}\\
MOD \pub{NeurIPS 24} \cite{MOD}          &N  &N  &0  &\checkmark &\checkmark &\checkmark &\checkmark\\
Args \pub{ICLR 24} \cite{Args}         &0  &\textgreater N &0  &\checkmark &\ding{55} &\checkmark &\checkmark\\
Steering \pub{EACL 24}  \cite{steering}  &0  &1  &0  &\ding{55}  &\ding{55} &\checkmark &\checkmark\\
MetaAligner \pub{NeurIPS 24} \cite{MetaAligner}  &1  &2  &1  &\ding{55 } &\ding{55}  &\checkmark &\ding{55}\\
LoraMoE \pub{ACL 24} \cite{LoraMoE}      &1  &1  &1 &\ding{55} &\checkmark &Retrain &\checkmark\\
PCB-Merging \pub{NeurIPS 24} \cite{PCBmerging}  &1  &1  &0  &\ding{55} &\checkmark &\checkmark &\checkmark\\
PAD \pub{ICLR 25} \cite{PAD}          &1  &3  &1  &\ding{55}  &\ding{55} &Extra-train &\ding{55}\\
GenARM \pub{ICLR 25} \cite{GenARM}          &0  &\textgreater N &0  &\checkmark &\ding{55} &\checkmark &\checkmark\\
RARM \pub{ICML 25} \cite{RARM}          &0  &2  &1  &\checkmark &\ding{55} &Retrain &\checkmark\\     
\cellcolor[HTML]{FDF5F1}\textbf{\ourapproach (ours)}        &\cellcolor[HTML]{FDF5F1} 1  &\cellcolor[HTML]{FDF5F1} 1  &\cellcolor[HTML]{FDF5F1} 0  &\cellcolor[HTML]{FDF5F1} \checkmark &\cellcolor[HTML]{FDF5F1} \checkmark &\cellcolor[HTML]{FDF5F1} \checkmark  &\cellcolor[HTML]{FDF5F1} \checkmark \\
\hline 
\end{tabular}
}
\label{table:comparison}
    \vspace{-3pt}
\end{table*}
 The main contributions of this study are as follows:
 \begin{itemize}
\setlength{\itemindent}{-5mm}
 \item
 We investigate a novel decomposition strategy that breaks down the multi-objective alignment problem into a series of single-preference subproblems, each handled by a set of specialized experts, enabling fine-grained control and full Pareto coverage.
 \item 
  We propose \ourapproach, a \textit{lightweight}, \textit{parameter-efficient} and \textit{plug-and-play} hierarchical Mixture-of-Experts framework that comprises three-level hierarchy, bypassing full model training.
 \item 
 We evaluate \ourapproach across diverse multi-, many-objective and multi-task settings, involving 14 objectives, 6 benchmarks, and 200 preference. \ourapproach consistently outperforms 15 recent baselines with  lower training cost and parameter overhead.
\end{itemize}

\section{Related Work}
\paragraph{LLM Multi-objective Alignment.}
MORLHF~\cite{MORLHF} and MODPO~\cite{MODPO} employ linear scalarization to combine multiple reward signals into a single scalar metric, applying standard RLHF or DPO training a separate model for each preference.
Multi-objective Decoding (MOD)~\cite{MOD}, Alignment as Reward-Guided Search (Args)~\cite{Args} and Personalized Alignment at Decoding-time (PAD)~\cite{PAD} derive a closed form solution of optimal preference model and perform linear fusion of logits prediction during decoding. 
Directional Preference Alignment (DPA)~\cite{DPA} and Reward-in-Context (RiC)~\cite{RiC} typically inject user preferences into the prompt, enabling in-context preference-conditioned alignment.
Additionally, Steering~\cite{steering, steering0} adding ``steering vectors'' to all token positions after the user's prompt, enabling precise control over multi-objective preferences. 
MetaAligner~\cite{MetaAligner} extends the Aligner~\cite{Aligner} framework to MOA, refining weaker outputs to better match user preferences.
\vspace{-10pt}
\paragraph{Knowledge Fusion for LLMs.} 
Model merging~\cite{RegMerging,FisherMerging,PCBmerging, FREE-Merging,Ties-merging,guodong24acl,guodong_acl25,junlin_acl25} is a widely used fusion technique that integrates multiple task-specific models into a unified model.
Task Arithmetic (TA)~\cite{TaskArithmeic,RS,PersonalizedSoups} linearly combines task vectors, defined as the parameter differences between task-specific models and the original pre-trained model. 
Then RS~\cite{RS} and PS ~\cite{PersonalizedSoups} firstly extend this concept to MOA.
LoraMoE~\cite{LoraMoE}, the closest work to ours, is a Mixture-of-Experts (MoE) approach that uses LoRA Adapters~\cite{LoRA} as experts, integrating LLM knowledge by activating select experts via a router network.
However, it requires costly training across all LoRA experts simultaneously and limits knowledge sharing among them, thus unsuitable for MOA.

In summary, we systematically compare and analyze existing LLM alignment methods in Tab.~\ref{table:comparison}.

\begin{figure*}[t]
    \centering \includegraphics[width=0.99\linewidth, trim=0 0 0 35, clip]{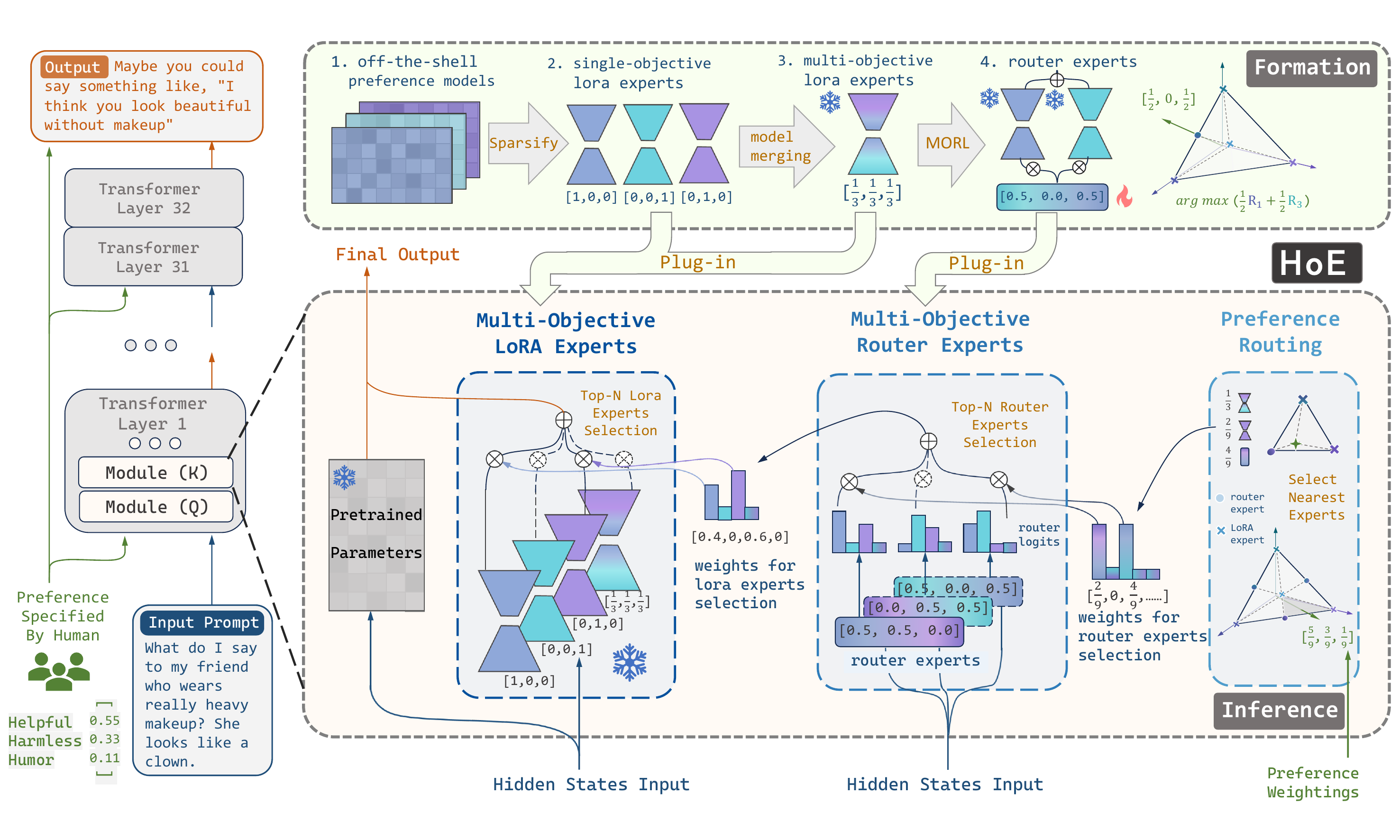}
    \vspace{0pt}
    \caption{Illustration of our \ourapproach approach. The left side illustrates the application scenario, where the model generates a response aligned with the prompt and given preferences. The bottom-right highlights its three \textit{\textbf{hierarchical}} components - the LoRA experts, router experts, and a preference routing.
    The top-right depicts individual components, each serving as an expert for specific weightings, designed for seamless plug-and-play integration within the model.}
    \label{fig:main architecture}
    \vspace{-10pt}
\end{figure*}
\section{Methodology}
In this section, we present the methodology behind \ourapproach, a lightweight, parameter-efficient, and plug-and-play multi-objective alignment framework.
As illustrated in Figure~\ref{fig:main architecture}, our \ourapproach approach consists of three \textit{hierarchical} components: LoRA experts, router experts and a preference routing.

\subsection{\texttt{Primary} LoRA Experts}
\label{text: MO Lora Experts}
\paragraph{Single-Objective LoRA Experts.}
 We begin with $N$ objectives $\{R_1,...,R_N\}$, a pre-trained model $\pi_{pre}$, and a collection of off-the-shell single-objective optimal policies $\{\pi^{*}_1,...,\pi^{*}_N\}$, each fine-tuned on its respective objective.
 \begin{align}
\pi^{*}_i  = \mathop{\arg\max}\limits_{\pi_\theta} \mathbb{E}_{x\sim D,  y\sim\pi_\theta(\cdot|x)}[R_i(x,y) - \beta \mathbb{KL}(\pi_\theta||\pi_{pre})]
\end{align}
Each model $\pi^{*}_i$ is initialized from pre-trained weights $\theta_{pre}$, and fine-tuned to obtain parameters $\theta_i$.

Following the task vector paradigm in model merging~\cite{TaskArithmeic}, we define each  ``objective vector'' as the difference between fine-tuned weights $\theta_i$ and the pre-trained weights $\theta_{pre}$: 
\begin{align}
\tau_i  = \theta_i - \theta_{pre}
\end{align} 
These extracted objective vectors inherently capture the single-objective capabilities of each single-objective model.

Then, we apply Task-Vector Singular Value Decomposition (task-SVD) to
obtain compact LoRA adapters for each objective: $A_i, B_i = \text{task-SVD}(\tau_i)$, where high-magnitude components of $\tau_i$ are extracted, followed by SVD and parameter rescaling to convert into new LoRA matrices $A_i$ and $B_i$.
These compact adapters, referred to as LoRA Experts, are highly specialized for their corresponding objectives, preserving optimal performance at the corresponding ends (\ie one-hot preference) of the Pareto frontier.
Empirical results ~\cite{Delta_Compression0,Delta_Compression1,Delta_Compression2,Delta_Compression3,Delta_Compression4} demonstrated its effectiveness with negligible performance loss across diverse LLMs and alignment objectives.

Consequently, we replace all linear module in the Transformer architecture with MoE-style plugin modules, incorporating the LoRA experts.
When receiving user preference $ \lambda \in \mathbb{R}^N$, we simply linearly combining the $N$ LoRA experts' outputs with weightings $ \lambda$, yielding:
\begin{align}
O_\lambda(x) =  W_{pre}x + \sum_{i=1}^N \lambda_i(B_iA_ix)
\end{align} \label{eq: linear combine single-O Lora experts}
where $x \sim \mathbb{R}^{d_{in}}, W_{pre}\sim \mathbb{R}^{d_{in}\times d_{out}}, A_i \sim \mathbb{R}^{d_{in}\times r},  B_i \sim \mathbb{R}^{r\times d_{out}}$ and rank $r \ll min(d_{in}, d_{out})$.

\paragraph{Multi-Objective LoRA Experts.}
For preferences involving multiple objectives simultaneously , the aforementioned linear combination of single-objective experts may fail to recover optimal performance, especially at intermediate points on the Pareto frontier (\eg $\lambda = [0.5, 0.5]$).
To address this, we draw inspiration from model merging~\cite{AdaMerging, FisherMerging, FREE-Merging,PCBmerging,RegMerging,Ties-merging} which amplify parameters beneficial to all tasks while suppressing conflicting or detrimental ones, enable nonlinear and fine-grained parameter adaptation, and significantly outperform linear approaches (\eg Task Arithmetic ~\cite{TaskArithmeic}).


To cover the entire Pareto frontier, we incorporate model merging into our framework to derive new expert parameters tailored to arbitrary preference vectors.
Given a target preference $\lambda$, we specify the desired objective proportions and synthesize a merged expert with parameters:
\begin{align}
\tau_\lambda = \text{Merge} (\{\tau_i\}_{i \in [N]}, \lambda)
\end{align}
where $\{\tau_i\}$ are the objective vectors derived from single-objective model.
We then reuse the same task-SVD procedure.
These resulting adapters serve as multi-objective LoRA experts, and are no longer aligned with a single objective, but instead specialized in specific combinations of objectives (\eg $[0.5, 0.5]$).

\subsection{\texttt{Secondary} Router Experts}
\label{text: MO Router Experts}

While increasing the number of LoRA experts improves performance, their number is limited by parameter overhead—since LoRA adapters are not negligible in size.
To address these challenges, we propose a lightweight and fine-grained decomposition method called Router Experts, which introduces one-layer linear routers as secondary experts.
The parameter size of such a component is negligible compared to the LoRA experts, while it plays a crucial role in enhancing the results, due to the static nature of router: \texttt{module-wise fine-grained routing}, and \texttt{input-adaptive selection}.
This enables our router experts to dynamically select the most suitable LoRA experts based on the input, allowing for module-wise and more efficient utilization of LoRA parameters —a key distinction from LoRA experts.

\paragraph{Formation.}
Each router $r_\lambda$, represented by the weighting $\lambda$, consists of different router layers distributed across all Transformer modules.
A router layer is a linear network $W \in \mathbb{R}^{d_{in}\times N}$ and $ b \in \mathbb{R}^N$, taking the same input (hidden states $x$) as the LoRA adapters, producing scores $W^Tx +b$.
Notably, each router expert only votes for the $N$ nearest LoRA experts in its simplex region (detailed in \ref{text: Integration of All Components}).
As each router expert is optimized with respect to a specific weighting $\lambda$, it qualifies as an expert tailored to that particular preference.

\paragraph{Optimization.}
Unlike LoRAMoE~\cite{LoraMoE}, our method keeps LoRA expert parameters frozen, drastically reducing training resource requirements. 
Each router expert $r_\lambda$ is optimized for its corresponding weighting $\lambda$ with the frozen LoRA experts, making them \texttt{lightweight, plug-and-play} modules.
Given  L existing LoRA experts $\{\tau_{\lambda_i}\}_{\lambda_i\in \Lambda}$, each router activates only the $N$ nearest neighbor to $\lambda$.
The training objective is to maximize the linearly-combined reward $R_\lambda(\cdot) = \sum_i \lambda_i R_i(\cdot) $ under a mixture policy composed of these selected experts:
\begin{align}
r_\lambda =  \mathop{\arg\max}\limits_{r} \ \mathbb{E}_{y\sim\pi_r(\cdot |x)} [R_\lambda(x,y)], s.t. \pi =  HoE(\theta_{pre}, \tau_{\mathbb{NN}_N(\lambda, \Lambda)})
\end{align}
To capture non-convex regions of the Pareto Front, we employ Tchebycheff (TCH) scalarization, optimizing for the worst-case objective:
\begin{align}
     \mathbb{J}(\theta | \lambda) = \mathop{max}\limits_{\theta} \mathop{min}\limits_{i} \{\lambda_i(R_i(x,y) - z^{*}_i)\}
\end{align}
where $z^{*} \in \mathbb{R}^N$ is a reference point , and $\lambda$ now encodes objective importance, and $\mathbb{E}_{x\sim D, \ y\sim\pi_\theta(\cdot |x)}[R_i(x,y)]$ is abbreviated as $R_i(\theta)$.
We solve this max-min objective via Online Mirror Descent (OMD)~\cite{OMD}, yielding:
\begin{align}
    \mathbb{J}(\theta|\lambda) = \mathop{max}\limits_{\theta} \sum_i \{w_i (R_i(\theta) - z^{*}_i)\}  \text{\ s.t.\ }  w = \mathop{min}\limits_{w} \{w_i \lambda_i (R_i(\theta) - z^{*}_i)\} \text{\ \ and \ \ } {||w||}_1 = 1  \label{eq:TCH}
\end{align} 
The indicator vector $w$ is smoothed ~\cite{STCH} and updated using TD-learning ~\cite{OMD-STCH-MORL}  for stability.
We seamlessly integrate this ideal process into the PPO~\cite{schulman2017proximal} paradigm, resulting in a mixed-advantage formulation:
\begin{align} \label{equa: Policy Gradient MORL}
    \nabla_\theta\mathbb{J}(\theta|\lambda) =   \mathbb{E}_{s_t,a_t \sim \pi(s_{t-1})}[ \big( \sum_{i=1}^N w_i A_i^{\pi_\theta}(s_t,a_t)\big) \nabla_\theta log \pi_\theta (s_t,a_t)]  
\end{align}
More details of the practical implementation see Appendix \ref{Appendix: Implementation Details}.
\ourapproach enjoys a theoretical convergence rate of $O( log\frac{N}{T})$ where N is the number of objectives and T is the number of iteration rounds (see Appendix~\ref{Appendix: Proof}).

\subsection{\texttt{Tertiary} Preference Routing}
The preference routing module is parameter-free layer based on geometric proximity, mapping a user's continuous preference vector $\lambda_{user}$ to a discrete subset of experts. 
As illustrated  in Fig.~\ref{fig:main architecture}, preference routing module partitions the entire $N$-simplex into coarse regions defined by LoRA experts, and further subdivide them into finer subregions via router experts. 
This hierarchical decomposition enables finer-grained $N$-objective alignment for any user's preference.
Formally, it selects the $N$ closest expert weightings from set of all M expert $\Lambda = \{\lambda_i\}_{i \in [M]}$ based on Euclidean distance:
\begin{align}
 \mathbb{NN}_N(\lambda_{user}, \Lambda) = \mathop{\arg\min}\limits_{i}^{N} \|\lambda_{user} - \lambda_i\|
\end{align} 
This mapping assigns $\lambda_{user}$ to a corresponding subregion spanned by the selected $N$ experts, guiding downstream router expert activation.


\subsection{Hierarchical Assembly for Inference}
\label{text: Integration of All Components}
\vspace{-2pt}
At inference time, we assemble the three layers into a unified hierarchical model that maps a user’s preference vector $\lambda_{user}$ to a tailored expert composition for response generation. 
We begin with a collection of  $L$ LoRA experts and $R$ router experts $\Theta = {\{\tau_i\}}_{i\in [L+R]}\ , \Lambda = {\{\lambda_i\}}_{i\in [L+R]}$.

\paragraph{Preference Routing.}
Tertiary preference routing module selects the N nearest experts and then expresses $\lambda_{user}$ as a convex combination of their preference vectors: 
\begin{align}
\omega_r\  \  \ s.t.\ \lambda_{user} = \omega_r^T[\lambda_i]_{i \in \mathbb{NN}_N(\lambda_{user}, \Lambda)}
\end{align}
The resulting voting vector $\omega_r \in \mathbb{R}^{L+R}$ serves as a soft selector over router experts, where only the selected top-N entries are nonzero.

\paragraph{Router Expert Voting.}
Each router expert $r_{\lambda_i}$ produces routing logits based on the input hidden states.
The final router output is obtained by aggregating  their predictions, weighted by  $\omega_r$: 
\begin{align}
    \omega_l = \omega_r^T[r_{\lambda_i}(x)]_{i \in [R]}
\end{align}
This weighted combination determines the activation of LoRA experts.

\paragraph{LoRA Expert Composition.}
Finally, the transformer’s output is computed as a mixture of the selected LoRA experts, combined with the pre-trained base weights:
\begin{align}
O_\lambda(x) =  W_{pre}x + \sum_{i=1}^N (\omega_{l,i}B_iA_ix) 
\end{align}
\vspace{-18pt}
\section{Experimental Setup}
\textbf{Datasets and Objectives}. 
We follow prior multi-objective alignment studies~\cite{MOD, RiC, RS, PAD}, using seven text generation tasks—Helpful Assistant, Math, Reddit Summary, Beaver Tail, Helpsteer, Psoups, and Helpsteer2 — covering 14 objectives. 
For more details, refer to Appendix~\ref{Appendix: Experiment Details}.

\textbf{Baselines.}
We consider 15 competitive algorithms as baselines: RS~\cite{RS}, MOD~\cite{MOD}, MODPO~\cite{MODPO}, RiC~\cite{RiC}, MetaAligner~\cite{MetaAligner}, PAD~\cite{PAD}, MORLHF~\cite{MORLHF}, Args~\cite{Args}, Steering~\cite{steering}, LoraMOE~\cite{LoraMoE}, PCB-Merging~\cite{PCBmerging}, FR-Merging~\cite{FREE-Merging}, Aligner~\cite{Aligner}, Preference-prompting and Personalized soups~\cite{PersonalizedSoups}. 

\textbf{Metrics.} 
We primarily use reward model scores to obtain the Pareto frontiers (Fig.\ref{fig:main_PF})—each objective is paired with a commonly used open-source reward model. Additionally, we report GPT-4-based win rates—comparative against base model—for further evaluation. Specially, for the math benchmark, we report PASS@1 accuracy, and for the over-refusal benchmark~\cite{OR-bench}, we report safety and helpfulness ratio. For more details, refer to Appendix~\ref{Appendix: Experiment Details}.


\section{Main Results}
\begin{figure*}[t]
    \centering
\includegraphics[width=1.0\linewidth,trim=3 0 5 0, clip]{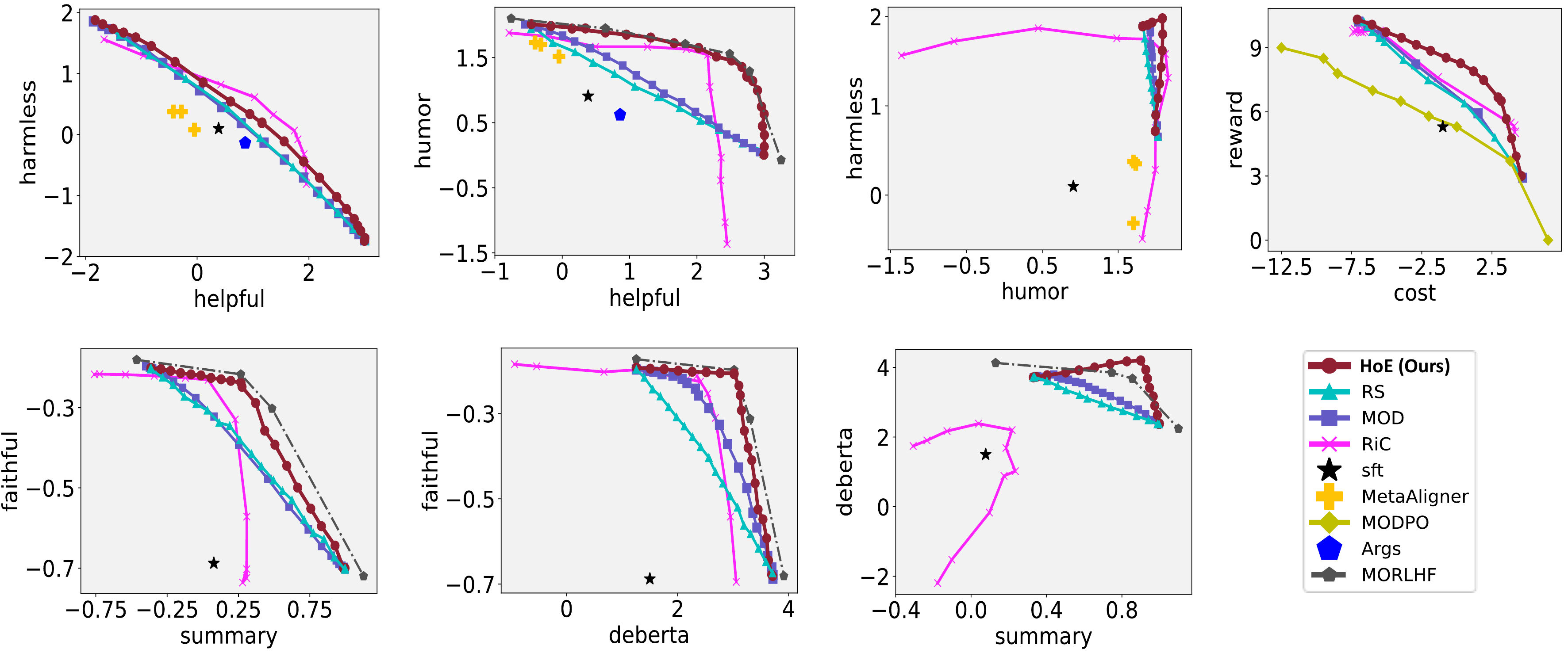}
    \vspace{0pt}
    \caption{Results of two-objective alignment on HelpAssistant, Reddit Summary and BeaverTails Task with 8 objectives. Compared to the baselines, \ourapproach consistently achieves superior Pareto frontiers.}
    \label{fig:main_PF}
    \vspace{-10pt}
\end{figure*}

\begin{figure*}[ht]
    \centering
    \vspace{-10pt}
    \includegraphics[width=1.0\linewidth,trim=120 10 105 40, clip]{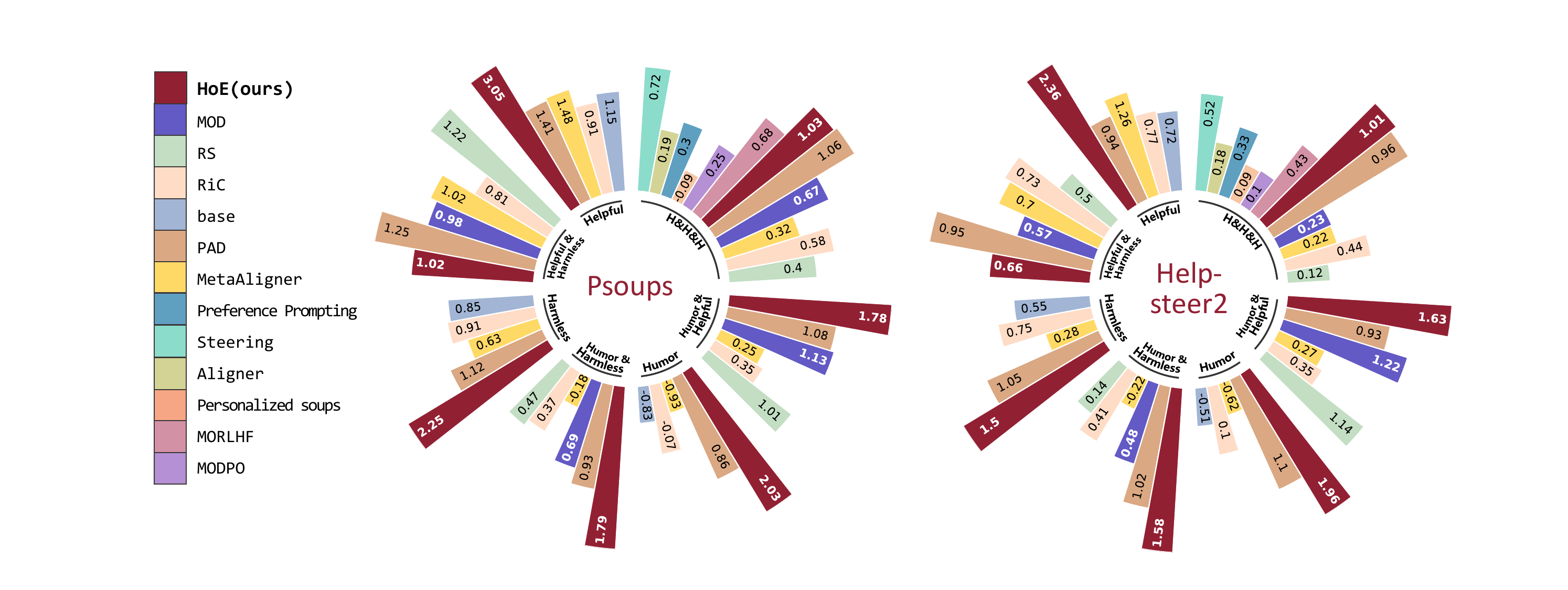}
    \vspace{-15pt}
    \caption{Comparison of alignment results with three objectives (\ie helpful, harless and humor) on the Psoups and Helpsteer2 datasets. }
    \label{fig:barplot}
    \vspace{-5pt}
\end{figure*}

We conduct experiments on 6 different NLP tasks with 14 different objectives, testing 200 different preferences and comparing them with 15 baselines. 
Experiments span two-, three-, and many-objective alignment scenarios. Quantitative comparisons are shown in Fig.\ref{fig:main_PF}, Fig.\ref{fig:barplot}, and Tab.~\ref{table: five obj result}.
\subsection{Two-Objective Alignment Results}

Fig.~\ref{fig:main_PF} presents the results for two-objective alignment across seven setups. \ourapproach clearly approaches the theoretical upper bound defined by MORLHF, producing smooth and convex Pareto frontiers, strongly validating its effectiveness.

In all cases, \ourapproach clearly outperforms RS and MOD—our Pareto frontier fully dominates theirs across all preference weightings. 
Even when constrained to use only LoRA experts for fairness, our method retains this dominance (see Fig.\ref{fig:ablation}).

Compared to RiC, \ourapproach achieves better results in 5 out of 7 cases. In the ``Summary \& Deberta'' setting, for instance, our model outperforms RiC by a notable $(+2, +0.8)$ margin. 
Although RiC slightly outperforms us in a few specific weightings (\eg ``Helpful \& Harmless''), this is likely due to its advantage in handling strongly conflicting objectives via online training.


Meanwhile, MetaAligner and Args are limited to the Helpful Assistant task, where their performance is comparatively weak.
MODPO also falls significantly short on the BeaverTail task, indicating better generalization of \ourapproach.
\subsection{Three-Objective Alignment Results}
We evaluate alignment across three objectives—Helpful, Harmless, and Humor—on the Helpful Assistant task (see Fig.~\ref{fig:3obj}). \ourapproach Pareto-dominates RS and MOD, and consistently outperforms RiC across most of the weight space.


We further test on Psoups and HelpSteer2 using LLaMA3.1-8B, comparing with 11 baselines under a strict generalization setting (none of the models were trained on these datasets).
As shown in Fig.~\ref{fig:barplot}, our method ranks first in 11 out of 14 evaluation setups. In the remaining three, PAD slightly outperforms us—yet we remain highly competitive.

Additionally, GPT-4-based evaluations (see Appendix.~\ref{Appendix: addtional exper result} and Fig.~\ref{fig:GPT4}) align closely with reward model scores, further confirming the robustness of our approach across models and tasks.
\begin{wrapfigure}{r}{0.65\textwidth}
\setlength{\tabcolsep}{2pt}
\vspace{30pt}
\caption{Five-objective alignment results on HelpSteer. Preference weighting settings are shown in gray. The best results are bolded and second best ones are underlined.}
\label{table: five obj result}
\vspace{-10pt}
\begin{center}
\begin{small}
\begin{sc}
  \resizebox{0.65\textwidth}{!}{
  \begin{tabular}[width=\columnwidth]{l|ccccc|c}
    \toprule
    \textbf{Method} & \textbf{Helpful} & \textbf{Correct} & \textbf{Coherence} & \textbf{Complex} & \textbf{Verbosity} & \textbf{Average} \\
    \hline
    \cellcolor[HTML]{E8E8E8}preference & \cellcolor[HTML]{E8E8E8}0.2 & \cellcolor[HTML]{E8E8E8}0.2 & \cellcolor[HTML]{E8E8E8}0.2 & \cellcolor[HTML]{E8E8E8}0.2  & \cellcolor[HTML]{E8E8E8}0.2  & \cellcolor[HTML]{E8E8E8} \\
RS              & 67.2 & 68	  & 76.8 & 37.3	 & 41.9  & 58.24\\
RiC             &\textbf{71.5}	&\underline{70.7}	&\textbf{78.3}	&\underline{41.1}	&43.8	&61.08 \\
MOD             & 68.4	&69.1	&76.6	&40	&\underline{45.9}	&60 \\
\cellcolor[HTML]{FDF5F1}\textbf{\ourapproach} (ours)  & \cellcolor[HTML]{FDF5F1}\underline{70.4} & \cellcolor[HTML]{FDF5F1}\textbf{71.6} & \cellcolor[HTML]{FDF5F1} \underline{78.1} & \cellcolor[HTML]{FDF5F1}\textbf{42.8}  & \cellcolor[HTML]{FDF5F1}\textbf{47.5}  & \cellcolor[HTML]{FDF5F1}\textbf{62.1 }\textcolor{cyan}{(+3.8)}		
\\\hline
\cellcolor[HTML]{E8E8E8}preference & \cellcolor[HTML]{E8E8E8}0.17 & \cellcolor[HTML]{E8E8E8}0.17 & \cellcolor[HTML]{E8E8E8}0.17 & \cellcolor[HTML]{E8E8E8}0.25  & \cellcolor[HTML]{E8E8E8}0.25  & \cellcolor[HTML]{E8E8E8} \\
RS              & 66.7 & 67.8 &	76.2 & 38.9	 & 42.6 & 58.44\\
RiC             & \underline{70}	&67.6	&\underline{76.5}	&\underline{42.3}	&46.2	&60.52 \\
MOD             & 68.1	&\underline{68.9}	&76.3	&40.9	&\underline{47.1}	&60.26 \\
\cellcolor[HTML]{FDF5F1}\textbf{\ourapproach} (ours)  & \cellcolor[HTML]{FDF5F1}\textbf{70} & \cellcolor[HTML]{FDF5F1}\textbf{71.1} & \cellcolor[HTML]{FDF5F1}\textbf{77.7} & \cellcolor[HTML]{FDF5F1}\textbf{42.9}  & \cellcolor[HTML]{FDF5F1}\textbf{48.7}  & \cellcolor[HTML]{FDF5F1}\textbf{62.08}\textcolor{cyan}{(+3.6)}					
\\\hline
\cellcolor[HTML]{E8E8E8}preference & \cellcolor[HTML]{E8E8E8}0.11 & \cellcolor[HTML]{E8E8E8}0.11 & \cellcolor[HTML]{E8E8E8}0.11 & \cellcolor[HTML]{E8E8E8}0.33  & \cellcolor[HTML]{E8E8E8}0.33  & \cellcolor[HTML]{E8E8E8} \\
RS              & 66.4 & 67.5 & \underline{75.8} & 40.5	& 44.3 & 58.9 \\
RiC             & \underline{67.7}	&62.4	&73.9	&\textbf{44}	&\textbf{49.9}	&59.58 \\
MOD             & 67.7	&\underline{68.2}	&75.6	&42.9	&48.1	&60.5 \\
\cellcolor[HTML]{FDF5F1}\textbf{\ourapproach} (ours)  & \cellcolor[HTML]{FDF5F1}\textbf{69.8} & \cellcolor[HTML]{FDF5F1}\textbf{70.8} & \cellcolor[HTML]{FDF5F1}\textbf{77.4} & \cellcolor[HTML]{FDF5F1} \underline{43.2}  & \cellcolor[HTML]{FDF5F1} \underline{49.3}  & \cellcolor[HTML]{FDF5F1}\textbf{62.1 }\textcolor{cyan}{(+3.2)} 				
\\\bottomrule
      \end{tabular}}
    \label{appendixtab:methods_evaluation}
\end{sc}
\end{small}
\end{center}
\vspace{-30pt}
\end{wrapfigure}



\subsection{Many-Objective Alignment Results}
We evaluate five-objective alignment on HelpSteer, with results presented in Tab.~\ref{table: five obj result}.
The \textsc{Preference} column indicates the user's preference vector $\lambda_{user}$. 
\ourapproach achieves the highest average score, outperforming MOD and RiC across all objectives, with only slight underperformance on a few specific objectives compared to RiC. This demonstrates that \ourapproach is highly effective for many-objective alignment.

\section{Analysis}
\subsection{Ablation Study}
\begin{figure*}[t]
    \centering
    \includegraphics[width=1.0\linewidth,trim=0 25 0 200, clip]{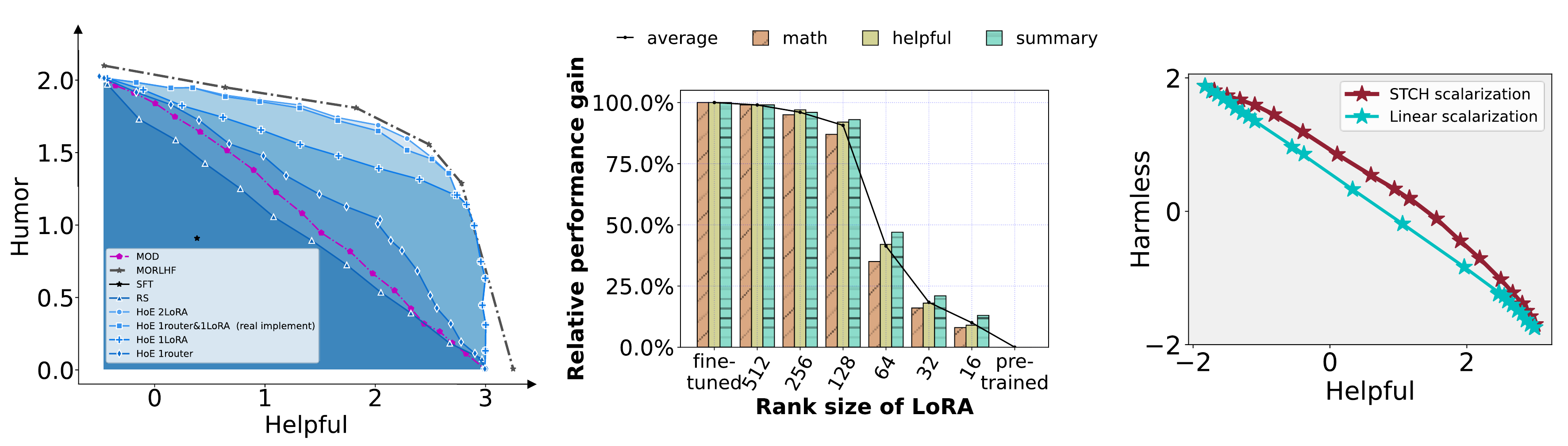}
    \vspace{-10pt}
    \caption{
    Ablation studies assessing the impact of expert count (Left), LoRA ranks (Middle), and Tchebyscheff scalarization (Right).}
    \label{fig:ablation}
    \vspace{-10pt}
\end{figure*}

We conducted three ablation studies to assess the impact of (1) individual expert, (2) LoRA ranks, and (3) Tschebyscheff scalarization:
\vspace{-10pt}
\paragraph{Ablation on Experts.}
This is the core ablation of our work. 
We isolate the roles of each LoRA experts and router experts by incrementally removing or combining them to observe their effect on the Pareto frontier (PF). 
All configurations in this study include two fixed single-objective LoRA experts, and the terms ``1 LoRA'' or ``2 LoRA'' refer specifically to the additional ones.
Fig.~\ref{fig:ablation} (left) presents results across four configurations:

1) \textit{\underline{1 Router}}: 
Adding a single router expert improves performance on specific preferences, highlighting its specialization capability. However, due to its smaller parameter count, the improvements are modest.
2) \textit{\underline{1 LoRA}}: 
A single LoRA expert leads to substantial PF expansion near its preference. But for other preference settings, performance drops off, revealing limited coverage.
3) \textit{\underline{1 LoRA \& 1 Router}}: 
This combination achieves a near-complete PF. The router complements the LoRA expert by covering underrepresented regions, showcasing their strong synergy.
4) \textit{\underline{2 LoRA}}:
Adding a second LoRA expert improves notable performance-close to the PF achieved by MORLHF, but with diminishing returns. In most cases, the 1 LoRA \& 1 Router setup offers good coverage with fewer parameters.
These results demonstrate that each expert plays a distinct role in shaping the PF, and that combining diverse expert types strike a balance between performance and parameter efficiency.
\vspace{-10pt}
\paragraph{Ablation on LoRA Rank.}
We investigate the effect of LoRA rank using LLaMA2-7B-Chat as the base model. For the math task, we extract Math LoRA experts from MathLLaMA2-7B; for assistant tasks, we use HelpfulLlama2-7B and HumorLlama2-7B.

Fig.~\ref{fig:ablation} (middle) shows that lower ranks lead to greater performance degradation. For math, ranks below 128 result in noticeable drops, suggesting high complexity. While assistant tasks are less challenging, and a rank size of 128 is sufficient.


\vspace{-10pt}
\paragraph{Ablation on Tschebyscheff Scalarization.}
We compare Tschebyscheff scalarization with linear scalarization in MORL. As shown in Fig.~\ref{fig:ablation} (right), linear scalarization often biases the policy to drift significantly toward PF edges, leading to instability or collapse. In contrast, Tschebyscheff-based optimization (OMD-STCH-MORL) maintains stable training while preserving full PF coverage. This confirms its advantage in multi-objective optimization stability.

\subsection{Advantages over Existing Methods}\label{txt: Comparison with others}
 While existing methods each excel in specific areas, \ourapproach offers seven notable advantages, with quantitative comparisons provided in Tab.~\ref{table: comparison param}.
 The checklist of advantages are listed in Tab.\ref{table:comparison}.

 1) \textit{Lightweight and Parameter-Efficient}. 
 All preference models are unified in a single architecture, significantly reducing storage demands, compared to methods that train and store multiple models.
 
 
 2)~\textit{Predominantly Training-free}. 
 \ourapproach relies primarily on model fusion, requiring minimal training only for a small portion of the router.
 While other methods (\eg RiC, PAD, and MetaAligner) require costly exhaustive training as objectives increase. 
 
 3)~\textit{Minimal Inference Cost}.
 \ourapproach activates only a few lightweight experts at inference time, making it much faster than decoding- or refinement-based methods (\eg MetaAligner, PAD, MOD) that require multi-pass inference.
 
 4) \textit{Applicable to Multi-task Learning}. 
 As demonstrated in Fig.~\ref{fig:MTL_radar}, \ourapproach achieves comparable performance to other baselines in multi-task learning scenarios, without specialized design for MTL.
 
 5) \textit{Pareto-Steerable.}
 \ourapproach supports arbitrary user preference, enabling continuous traversal along the Pareto frontier—unlike baselines fixed to preset preferences (\eg MetaAligner, PAD, and Steering).

 6) \textit{Plug-and-play and Scalable}. 
 New unseen objectives can be added without retraining existing experts; existing ones remain valid by simply extending the preference vector (\eg from $[0.5, 0.5]$ to $[0.5, 0.5, 0.0]$). 
 Some methods (\eg MORLHF, MOPPO, and RiC) require extensive retraining to involve the new objective, and others (\eg DPA, PAD, MetaAligner, and LoRAMoE) render previous checkpoints obsolete and necessitate complete retraining.
 
 7) \textit{Free from Prompting}.
 \ourapproach avoids reliance on handcrafted prompts, enabling generalization to abstract or hard-to-verbalize objectives (\eg ``deberta'', ``reward'' or ``cost'' in Fig.~\ref{fig:main_PF}) and preserving the core capabilities of the base LLM - unlike prompt-dependent methods(\eg PAD, MetaAligner, and DPA)

\section{Limitation and Future Work}
\label{Appendix: Limitation}
Despite the strengths of our method, several limitations remain:
(1) Our approach depends on off-the-shell single-objective models, which may not always be available. Training such models from scratch can be time-consuming and impractical in some settings.
(2) The method relies on effective model merging and SVD-based compression. 
While these techniques work well for the objectives considered, they may fail in some settings.

\section{Conclusion}
We propose \ourapproach, a hierarchical Mixture-of-Experts framework for multi-objective alignment in LLMs. By combining LoRA experts, router experts, and preference routing, our method enables efficient and scalable alignment across diverse user preferences. Experiments on 14 objectives across 6 benchmarks and 200 preferences show that \ourapproach outperforms 15+ strong baselines, achieving superior Pareto-optimal results in various multi-objective and multi-task settings.


\newpage
\section*{Acknowledgements}
This work was supported by National Science Foundation of China (62476070), Shenzhen Science and Technology Program \seqsplit{(JCYJ20241202123503005,~ GXWD20231128103232001,~ ZDSYS20230626091203008,~ KQTD2024072910215406)}  and Department of Science and Technology of Guangdong (2024A1515011540).
This work was also supported in part by the Major Key Project of PCL under Grant PCL2024A06 and PCL2022A05, and in part by the Shenzhen Science and Technology Program under Grant RCJC20231211085918010.

\bibliography{example_paper}


\appendix

\onecolumn



 
 
 
 

\section{The workflow of \ourapproach}
\label{Appendix C: Algorithms}
Algorithm~\ref{alg:pipline} show the whole pipeline of \ourapproach.


\begin{algorithm}[h]
   \caption{The workflow of \ourapproach}
   \label{alg:pipline}
\begin{algorithmic}
   \STATE {\bfseries Input:} objective number $N$, single-objective fine-tuned weights $\{\theta_i\}_{i\in [N]}$, pre-trained weights $\theta_{pre}$, N-Simplex $\Delta_N $, number of MO LoRA Experts $L$, number of MO router Experts $R$, \ourapproach model $\Theta = \{\}$
   \STATE uniformly select weightings $\{\lambda_l\}_{l\sim[N+L+R]} \sim \Delta_N $
   \FOR{$i=1$ {\bfseries to} $N$}
   \STATE $\tau_i \leftarrow $ extract LoRA from $(\theta_i - \theta_{pre})$
   \ENDFOR
   \FOR{$l=N$ {\bfseries to} $N+L$}
   \STATE $\tau_{i+l} \leftarrow$ Merging $\{\theta_i\}_{i\in [N]}$ with weighting $\lambda_l$
   \ENDFOR
   \STATE  $\Theta = \{\tau_i\}_{i=[N+L]}$
   \FOR{$r=N+L$ {\bfseries to} $N+L+R$}
   \STATE  $\tau_r \leftarrow$ Train router experts on $\lambda_r$ with $\Theta$
   \ENDFOR
   \STATE insert $\Theta = \{\tau_i\}_{i=[N+L+R]}$
   \OUTPUT $\Theta$
\end{algorithmic}
\end{algorithm}

\section{Additional Results}
\label{Appendix: addtional exper result}

\subsection{Mullti-Task Results}

\begin{figure}[ht]
    \centering
    \includegraphics[width=0.7\linewidth,trim=0 10 0 130, clip]{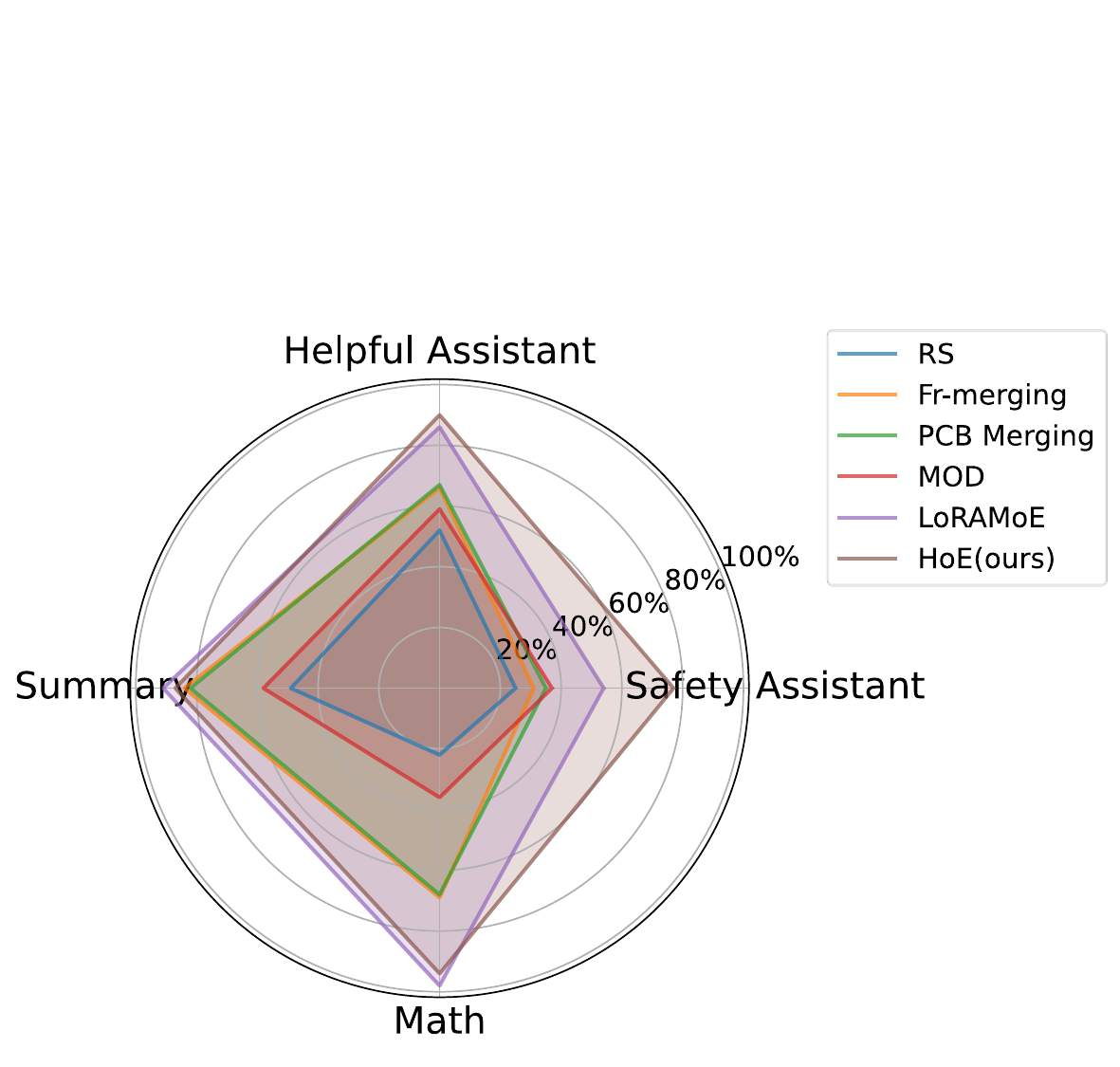}
    \vspace{0pt}
    \caption{Multi-Task Learning results. Our router experts specialized for ``Helpful Assistant'' and ``Safety Assistant'' enable better performance than LoRAMoE. The base model’s performance is normalized to 0\% and single-objective models are normalized to 100\%}
    \label{fig:MTL_radar}
    \vspace{0pt}
\end{figure}


We designed experiments involving four tasks learning:
1) Helpful Assistant: An assistant that provides helpful and correct responses to prompts, even for harmful ones.
2) Safety Assistant: An assistant that refuses to respond to harmful prompts.
3) Summary Task: Summarizes a given poster.
4) Math Task: Solves math problems from the GSM8K dataset\cite{GSM8K}.
The first two tasks were evaluated on the over-refusal benchmark\cite{OR-bench}, the Summary Task was assessed using the average score across three objectives, and the Math Task was evaluated with Pass@1 accuracy on the GSM8K test set. To balance different scores, all results were normalized, setting the base model’s performance to 0\% and single-objective models to 100\%, which are shown in Fig.\ref{fig:MTL_radar}.

We compared \ourapproach with baselines such as LoRAMoE, RS, MOD, PCBmerging, and FR-Merging, all initialized with the same model and using LoRA adapter-based fusion. As expected, \ourapproach outperforms PCBmerging, FR-Merging, and MoAlignment methods (e.g., RS, MOD). While LoRAMoE achieved strong performance on the Summary Task and Math Task, it struggled on the Helpful and Safety Assistant tasks due to the nuanced and overlapping nature of harmful and seemingly harmful prompts in the over-refusal benchmark. The router in LoRAMoE, designed for uniform preferences $[0.25,0.25,0.25,0.25]$, failed to distinguish between red-teaming prompts and less harmful ones effectively.
In contrast, \ourapproach introduced specialized router experts for the Helpful and Safety Assistant tasks ($[0.5,0.5,0.0,0.0]$), enabling better performance by dynamically adjusting input weightings. This improvement highlights the flexibility and robustness of \ourapproach in multi-task learning scenarios.




%


\setlength{\tabcolsep}{1pt}
\begin{table*}[h]
\caption{Alignment results for unseen dataset HelpSteer2 and Psoups on three objective s(\ie Helpful, Harmless, Humor) with Llama3.1-8B}
\centering
  \resizebox{0.6\textwidth}{!}{
  \begin{tabular}{lccccccc}
    \toprule
   \multirow{2}{*}{\textbf{Method}} & \multirow{2}{*}{\textbf{Helpful}} & \small\textbf{Helpful\&} & \multirow{2}{*}{\textbf{Harmless}} & \small\textbf{Harmless\&} & \multirow{2}{*}{\textbf{Humor}} & \small\textbf{Humor\&} &\multirow{2}{*}{\textbf{HHH}}\\
   & &\small\textbf{Harmless} & & \small\textbf{Humor}& & \small\textbf{Helpful}&\\
    \hline
    \multicolumn{8}{l}{\cellcolor[HTML]{E8E8E8}\textbf{Psoups dataset}}\\
Base            & 1.15 & 0.91 & 0.85 & -0.07 & -0.83 & 0.1  & 0.32 \\
RS              & 0.88 & 0.88 & 0.92 & 0.49  & 0.15  & 0.44 & 0.4 \\
RiC             & 0.91 & 0.81 & 0.91 & 0.37  & -0.07 & 0.35 & 0.58 \\
\small MetaAligner     & 1.48 & 1.02 & 0.63 & -0.18 & -0.93 & 0.25 & 0.32 \\
MOD             & 0.94 & 0.96 & 1.01 & 0.68  & 0.46  & 0.69 & 0.67 \\
PAD             & 1.41 & \textbf{1.25} & 1.12 & 0.93  & 0.86  & 1.08 & \textbf{1.06} \\
\cellcolor[HTML]{FDF5F1} \textbf{\ourapproach (Ours)}  & \cellcolor[HTML]{FDF5F1}\textbf{3.05} & \cellcolor[HTML]{FDF5F1}1.02 & \cellcolor[HTML]{FDF5F1}\textbf{2.25} & \cellcolor[HTML]{FDF5F1}\textbf{1.79}  & \cellcolor[HTML]{FDF5F1}\textbf{2.03}  & \cellcolor[HTML]{FDF5F1}\textbf{1.78} & \cellcolor[HTML]{FDF5F1}1.03

\\\hline
\multicolumn{8}{l}{\cellcolor[HTML]{E8E8E8}\textbf{HelpSteer dataset}}\\
Base            & 0.72 & 0.63 & 0.55 & -0.04 & -0.51 & 0.09 & 0.17 \\
RS              & 0.76 & 0.74 & 0.81 & 0.47  & 0.17  & 0.44 & 0.12 \\
RiC             & 0.77 & 0.73 & 0.75 & 0.41  & 0.10  & 0.35 & 0.44 \\
MetaAligner     & 1.26 & 0.70 & 0.28 & -0.22 & -0.62 & 0.27 & 0.22 \\
MOD             & 0.72 & 0.77 & 0.85 & 0.59  & 0.36  & 0.51 & 0.23 \\
PAD             & 0.94 & \textbf{0.95} & 1.05 & 1.02  & 1.10  & 0.93 & 0.96 \\
\cellcolor[HTML]{FDF5F1} \textbf{\ourapproach (Ours)} & \cellcolor[HTML]{FDF5F1}\textbf{2.36}  & \cellcolor[HTML]{FDF5F1}0.66 & \cellcolor[HTML]{FDF5F1}\textbf{1.5} & \cellcolor[HTML]{FDF5F1}\textbf{1.58} & \cellcolor[HTML]{FDF5F1}\textbf{1.96}  & \cellcolor[HTML]{FDF5F1}\textbf{1.63}  & \cellcolor[HTML]{FDF5F1}\textbf{1.01} \\\bottomrule
  \end{tabular}}
    \label{appendixtab:circle_barplot}
\end{table*}

\begin{figure*}[ht]
    \centering
    \includegraphics[width=0.7\linewidth,trim=0 0 1200 300, clip]{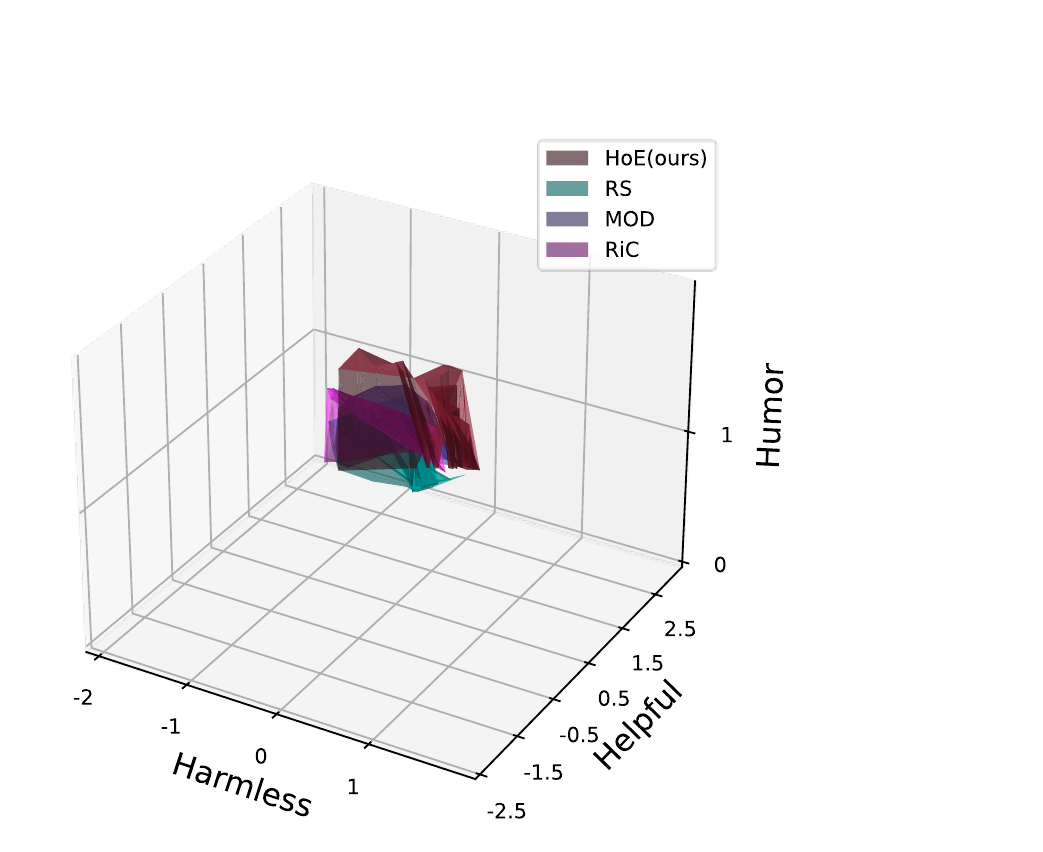}
    \vspace{-5pt}
    \caption{Alignment results with Helpful Assistant task on three-objective. Our approach consistently outperforms RS, MOD and RiC}
    \label{fig:3obj}
    \vspace{-10pt}
\end{figure*}
\newpage
\begin{figure*}[t]
    \centering
    \includegraphics[width=1.0\linewidth,trim=0 0 0 0, clip]{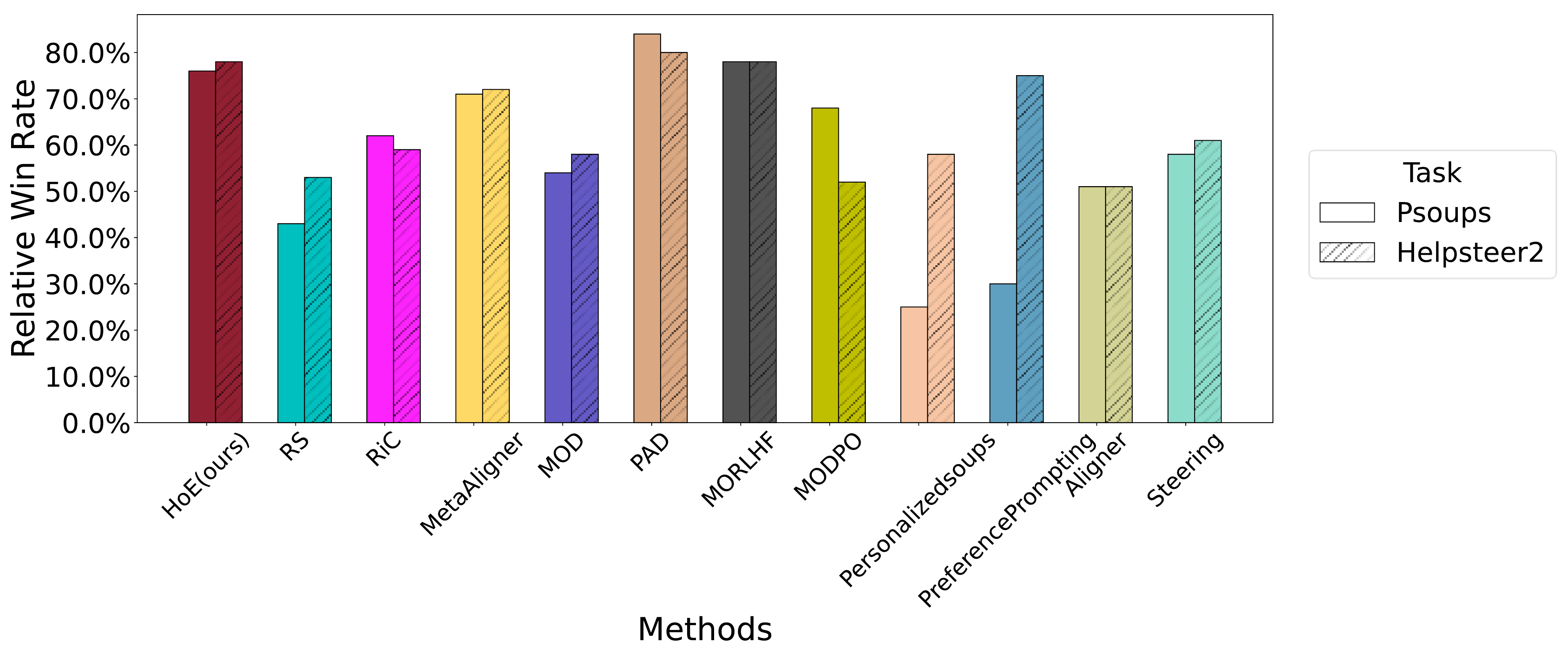}
    \vspace{-5pt}
    \caption{ GPT-4 evaluates all methods on Psoups and Helpsteer2 task, comparing the relative win rate of model-generated responses over the original responses for each approach. The evaluation is conducted across three dimensions: helpfulness, harmlessness, and humor. We take the average win rate across these three metrics as the final result.}
    \label{fig:GPT4}
    \vspace{-10pt}
\end{figure*}

\newpage
\subsection{Cost Analysis}

\begin{table}[h]
\setlength{\tabcolsep}{2pt}
\caption{Comparison of Training Costs, Storage Costs, and Inference Costs for various baselines, when using Llama2-7B as the base model to align on three objectives.}
\label{table: comparison param}
\begin{center}
\begin{small}
\begin{sc}
\begin{tabular}{lcccc}
\toprule
Baselines & Storage &  Training  & Inference \\
 &  & parameters &   cost \\
\midrule
RS   & 7.48B & 0 & 1.0\\
MOD & 7.48B & 0 &  $3.10\pm0.3$\\
MODPO    & 7.8B & 0.8B & 1.0\\
RiC    & 7.64B  & 0.64B &  1.0\\
Args     & 14B & 0.16B & $2.02\pm 0.2$\\
{\tiny MetaAligner}     &  14B & 0.16B & $2.04\pm 0.3$ \\
PAD      &   14B  & 0.16B & $2.98 \pm 0.5$\\
\ourapproach(ours)   & 7.64B & 8M & $1.23\pm0.2$ \\
\bottomrule
\end{tabular}
\end{sc}
\end{small}
\end{center}
\vskip -0.1in
\end{table}
We conduct a cost analysis of baseline models when performing three-objective alignment with LLaMA2-7B, as summarized in Table~\ref{table: comparison param}.
Our evaluation considers four key dimensions:
1)Storage: The amount of parameters that must be permanently stored in memory throughout the inference pipeline.
2)Number of Trainable Parameters,
3)Inference Cost: The computational overhead incurred during inference. 

Methods such as MetaAligner, Args, PAD, and MOD, which rely on decoding or refinement, significantly increase inference costs as the number of objectives grows. In contrast, \ourapproach only incurs a slight increase in inference time after activating three experts, demonstrating its scalability. Extrapolating from this, \ourapproach could align at least 12 objectives before inference time doubles, ensuring efficient multi-objective scaling.

Moreover, MetaAligner, Args, and PAD require at least two models at inference time. If full-parameter training is considered, PAD also requires storing an additional reference model, while MOD and RS each require three separate 7B-scale models. In contrast, \ourapproach extracts LoRA experts from full-rank dense task vectors and fine-tunes them to recover the optimal Pareto frontier, making it lightweight and highly parameter-efficient.

In terms of trainable parameters and training cost, \ourapproach requires significantly fewer parameters and resources than other training-based methods, making it a more efficient solution for multi-objective alignment.

\section{Potential Reader Questions}

\begin{table}[ht]
\setlength{\tabcolsep}{2pt}
\caption{Several methods against TA on three-objective alignment (Chinese \& Math \& Code)}
\label{table: MOexpert vs TA}
\begin{center}
\begin{small}
\begin{sc}
\begin{tabular}{lccc}
\toprule
&CMMLU	&GSM8K@1(5-shot)	&Human-Eval\\
\midrule
Pretrain	&-	&26.2	&-  \\
ChineseLlama	&38.6	&4.9	&13.4 \\
MathLlama	&31.2	&70.6	&0 \\
CodeLlama	&33.3	&28.0	&17.1 \\
TaskArithmetic\cite{TaskArithmeic}	&\underline{35.4}	&\underline{48.7}	&\underline{9.8} \\
PCB-Merging\cite{PCBmerging}	&36.5	&54.3	&16.5 \\
FR-Merging\cite{FREE-Merging}	&36.4	&55.6	&15.7 \\
TIES-Merging\cite{Ties-merging}	&36.4	&56.2	&14.0 \\
CodeExpert(r=256)	&-	&-	&16.7 \\
MathExpert(r=128)	&-	&66.3	&- \\
ChineseExpert(r=128)	&37.8	&-	&- \\
\textbf{MOLoRAExpert(ours)}	&\underline{35.7}	&\underline{50.4}	&\underline{13.7} \\
\bottomrule
\end{tabular}
\end{sc}
\end{small}
\end{center}
\vskip -0.1in
\end{table}

To further clarify several key design choices and theoretical intuitions, we address some potential questions that may arise when interpreting our method.

\textbf{Q1. Why is it necessary to merge the weights of different LoRA experts to construct additional experts, given that Eq.\ref{eq: linear combine single-O Lora experts} already performs a form of weight merging?}

One may question whether the weight merging in Eq.\ref{eq: linear combine single-O Lora experts} , which linearly combines LoRA experts, already suffices. While Eq.\ref{eq: linear combine single-O Lora experts}  indeed embodies a linear arithmetic operation akin to Task Arithmetic, this operation alone is insufficient to model the complex trade-offs required for multi-objective optimization. Our model merging strategy works at a finer granularity—directly at the parameter level—allowing selective reinforcement or attenuation of individual parameters. This more expressive mechanism enables us to better approximate solutions along the Pareto front. Empirical results (see Tab.~\ref{table: MOexpert vs TA}) merging improves performance by 40\%, and MOLoRA expert reducing storage by 30\% while retaining performance, confirming its necessity.

\textbf{Q2. Could other approaches such as MOD or RS similarly use LoRA to reduce storage?}

One may wonder whether alternatives like MOD and RS could benefit equally from LoRA-based compression. While both methods can, in theory, integrate LoRA to save storage, practical limitations arise.

In the case of RS, each LoRA adapter must be expanded into full-parameter form during inference, after which parameter soups are applied according to user preference. This results in storage requirements equivalent to full models and typically leads to inferior performance compared to directly using dense models.

MOD, on the other hand, can theoretically be adapted to use LoRA by applying different LoRA modules to a shared backbone. However, this design sacrifices one of MOD’s key strengths—cross-architecture decoding. Restricting MOD to a single base model severely limits its flexibility and practical deployment, making such an adaptation largely infeasible for real-world applications.

\textbf{Q3. Is there experimental evidence that Task Arithmetic (TA) underperforms in this context?}

One may ask for empirical evidence showing that Task Arithmetic yields subpar performance in our setting. Our ablation studies (see Fig.~\ref{fig:ablation}, Left) directly address this question. The results demonstrate that simply reducing the number of LoRA experts and performing naive arithmetic combinations significantly degrades performance, even falling behind MOD in some cases.

The only difference between TA and our “few-experts” configuration is that TA uses a fully parameterized vector while our method uses a sparse LoRA. To distinguish this, we have conducted ablation studies on LoRA ranks (see Fig.~\ref{fig:ablation}. Mid) and we further conducted experiments (see Tab.~\ref{table: MOexpert vs TA}) , showing TA underperforms other fusion methods, while LoRA Experts match TA with lower space cost.


\textbf{Q4. Does HoE need to use router experts to adaptively select LoRA experts during inference? Is there any analysis of the overhead?}

One may wonder whether router experts are necessary. Taking Llama3.1-8B as an example, the size of the router’s parameters is negligible compared to LoRA experts or the transformer's dense matrices, which we refer to as “parameter overhead.” Furthermore, traditional model merging struggles to handle extreme preference weightings (e.g., [0.1, 0.8, 0.1]), often leading to trivial MOLoRA experts (e.g., [0.33, 0.33, 0.33]). We refer to this issue as “coverage limitation.” 

While the parameters of router experts is negligible, its impact is far from negligible. During inference, router experts function as dynamic routers, enabling fine-grained selection of upper-layer LoRA experts. As shown in Fig.~\ref{fig:ablation} (Left), adding router experts can achieve a comparable effect to adding MOLoRA experts while maintaining lower parameter overhead.

\section{Implementation Details}
\label{Appendix: Implementation Details}
\subsection{LoRA Expert Details}
As the first work to apply task-SVD \cite{taskSVD} for sparsifying large-scale LLMs exceeding 7B parameters, one might reasonably question its effectiveness in this context. However, our experimental results demonstrate that model merging with task-SVD achieves remarkable performance in this domain.  

We now provide a detailed explanation of our sparsification process:  

Taking Math LoRA experts as an example, we compute the ``math task vector'' by subtracting LLaMA2-7B-Chat-hf from MathLLaMA-7B (an open-source model described in Appendix~\ref{Appendix: Experiment Details}). Next, we prune 40\% of the least significant parameters across the entire model based on absolute magnitude, resulting in a sparse matrix.  

We then apply SVD decomposition, sorting the singular values and selecting the top 128 rank-1 matrices. To further optimize performance, we dynamically determine a ``rescaling factor'' based on test results. In this case, we choose a scaling factor of $1.9$, which improves LLaMA2-7B-Chat-hf’s accuracy on GSM8K from $26\% \pm 4\% $to $68\% \pm 4\%$.




\subsection{\ourapproach Details}
For 2-objective alignment, in addition to the two corresponding single-objective LoRA experts, we introduce an additional LoRA expert represented by $[0.5, 0.5]$ and adaptively add one router expert based on the evaluation result. This results in a total of three LoRA experts and one router expert.

For 3-objective alignment, we include the three single-objective LoRA experts along with an additional LoRA expert represented by $[0.33, 0.33, 0.33]$. Specifically, for the Helpful Assistant Task, we incorporate a router expert represented by $[0.25, 0.25, 0.5]$ to enhance preference balancing. This results in a total of four LoRA experts and one router expert.

For 5-objective alignment on the HelpSteer Task, we utilize five single-objective LoRA experts alongside an additional LoRA expert represented by $[0.33, 0.33, 0.33, 0, 0]$ and a router expert represented by $[0.2, 0.2, 0.2, 0.2, 0.2]$ to improve adaptability across different preferences. This results in a total of six LoRA experts and one router expert.


\subsection{Optimization Details}
Unlike LoRAMoE~\cite{LoraMoE}, our method keeps LoRA expert parameters frozen, drastically reducing training resource requirements. 
Each router expert is optimized for its corresponding weighting with the frozen LoRA experts, making them plug-and-play and easy to integrate into existing architectures without extensive retraining.
The optimization objective for each router expert can be formalized as: 
\begin{align}
r_\lambda =  \mathop{\arg\min}\limits_{r} \ \mathbb{E}_{y\sim\pi_{HoE}(\cdot |x,\theta_{pre}, \tau_{1:N+E},  r)} [R_\lambda(x,y)]
\end{align}
where $\tau_{1:N+E}$ represents the set of existing LoRA experts, with only $N$ of them being activated for a given input.
$R_\lambda(x,y)$ is the linear combination of $N$ reward signals.
To capture non-convex regions of the Pareto Front, we employ Tchebycheff (TCH) scalarization, optimizing for the worst-case objective. 
For simplicity, $\mathbb{E}_{x\sim D, \ y\sim\pi_\theta(\cdot |x)}[R_i(x,y)]$ is abbreviated as $R_i(\theta)$.
We denote the expected reward for objective $i$ as $R_i(\theta) = \mathbb{E}_{x\sim D, \ y\sim\pi_\theta(\cdot |x)}[R_i(x,y)]$.
The objective can be formulated as: 
\begin{align}
     \mathbb{J}(\theta | \lambda) = \mathop{max}\limits_{\theta} \mathop{min}\limits_{i} \{\lambda_i(R_i(x,y) - z^{*}_i)\}
\end{align}
where $z^{*}$ represents a reference point indicating the desired performance level for each objective, and $\lambda$ denotes the relative importance of each objective. 

Due to the difficulty of directly solving the max-min formulation, we adopt the Online Mirror Descent (OMD) approach~\cite{OMD}. We decompose the TCH optimization into two stages:
\begin{align}
    \mathbb{J}(\theta|\lambda) = \mathop{max}\limits_{\theta} \sum_i \{w_i (R_i(\theta) - z^{*}_i)\}  \\
    \text{\ s.t.\ }  w = \mathop{min}\limits_{w} \{w_i \lambda_i (R_i(\theta) - z^{*}_i)\} \text{\ \ and \ \ } {||w||}_1 = 1  \label{eq:TCH2}
\end{align}
However, the one-hot nature of $w$ causes abrupt changes at Pareto frontier boundaries, affecting reinforcement learning robustness.
To address this, we use the Smooth Tchebycheff (STCH) approach~\cite{STCH}, replacing the min operator in Equation~\ref{eq:TCH2} with a softmax function, resulting in a smooth indicator vector 
$w$.
\begin{align}
     w = softmax \{ \lambda_i (z^{*}_i - R_i(\theta))\}
\end{align}
Following OMD-STCH-MORL \cite{OMD-STCH-MORL}, we update $w$ using TD-learning, enabling the optimization to leverage online data across multiple training batches for more stable estimation:
\begin{align}
    log \  w_i^{t+1} \leftarrow  log \  w_i^{t} + \alpha \lambda_i (z^{*}_i - R_i(\theta))
\end{align}
We seamlessly integrate the above ideal process into the PPO~\cite{schulman2017proximal} paradigm, resulting in a mixed-advantage formulation with the indicator vector $w$.
\begin{align} \label{equa: Policy Gradient MORL2}
    \nabla_\theta\mathbb{J}(\theta|\lambda) =   \mathbb{E}_{s_t,a_t \sim \pi(s_{t-1})}[A_\lambda^{\pi_\theta}(s_t,a_t) \nabla_\theta log \pi_\theta (s_t,a_t)]  \notag \\
    \text{where\ \ \ }A_\lambda^{\pi_\theta}(s_t,a_t) = \sum_{i=1}^N w_i A_i^{\pi_\theta}(s_t,a_t)
\end{align}


\section{Experiment Details}
\label{Appendix: Experiment Details}
\subsection{Datasets Details}

We utilize the following dataset for training and evaluation.

For Helpful Assistant task, we utilize ``hh-rlhf'' dataset \cite{hh-rlhf}\href{https://huggingface.co/datasets/Anthropic/hh-rlhf}{https://huggingface.co/datasets/Anthropic/hh-rlhf},  a multi-round dialogue dataset.

For Reddit Summary task, we utilize the summary dataset \cite{redditsummary} \url{https://huggingface.co/datasets/openai/summarize_from_feedback}.

For BeaverTails task, we utilize PKU-SafeRLHF-10K \cite{BeaverTails}\href{https://huggingface.co/datasets/PKU-Alignment/PKU-SafeRLHF-10K}{https://huggingface.co/datasets/PKU-Alignment/PKU-SafeRLHF-10K}

For HelpSteer task, we utilize the HelpSteer dataset.
\href{https://huggingface.co/datasets/nvidia/HelpSteer}{https://huggingface.co/datasets/nvidia/HelpSteer}

For Helpsteer2 task, we utilize the HelpSteer2 dataset
\href{https://huggingface.co/datasets/nvidia/HelpSteer2}{https://huggingface.co/datasets/nvidia/HelpSteer2}

For Psoups task, we utilize the same evaluation dataset as \cite{PersonalizedSoups}
\href{https://storage.googleapis.com/personalized-soups/data.zip}{https://storage.googleapis.com/personalized-soups/data.zip}

For Math task, we utilize the GSM8k dataset\cite{GSM8K} \href{https://huggingface.co/datasets/openai/gsm8k}{https://huggingface.co/datasets/openai/gsm8k}

For MTL task, we additionally utilize over-refusal benchmark \cite{OR-bench}.

\subsection{Reward Model Details}
The 14 distinct objectives consist of both interpretable natural language goals and names derived from reward models (RMs):
`` Helpful'', `` Harmless'', ``Humor'' on Helpful Assistant task, Psoups task and Helpsteer2 task; 
`` math'' on Math Task; 
`` faithful'', `` summary'', ``deberta'' on Reddit Summary task; 
`` reward'', `` cost'' on BeaverTail task;
`` helpfulness'', `` correctness'', `` coherence", `` complexity'', `` verbosity'' on Helpsteer task. 

We utilize following open-sourced reward models for training and evaluations. For Reddit Summary, we use \url{https://huggingface.co/Tristan/ gpt2_reward_summarization} for Summary, \url{https://huggingface.co/OpenAssistant/reward-model-deberta-v3-large-v2} for ``deberta'' and \url{https://huggingface.co/CogComp/bart-faithful-summary-detector} for Faithful; for Helpful Assistant, HelpSteer2 and Psoups, we use \url{https: //huggingface.co/Ray2333/gpt2-large-helpful-reward_model} for Helpfulness, \url{https://huggingface.co/Ray2333/gpt2-large-harmless-reward_model}  for Harmlessness and \url{https://huggingface.co/mohameddhiab/humor-no-humor}for Humor; for BeaverTail, we use \url{https://huggingface.co/PKU-Alignment/beaver-7b-v1.0-reward}
 for ``reward'' and \url{https://huggingface.co/PKU-Alignment/beaver-7b-v1.0-cost}
 for ``cost' for all five objectives, helpfulness, correctness, coherence, complexity and verbosity.

 The `` helpful'' and  `` harmless'' RMs are directly trained on `` hh-rlhf'' dataset with nearly 0.8 accuracy. The `` humor'' RM was trained on a joke dataset to detect humor with a 0.95 F1 score. The five RMs for HelpSteer are directly trained on HelpSteer with over 0.75 accuracy. 

\subsection{Base Model Details}
We utilize three base pre-trained models: Llama2-7B\cite{llama2-7b}\footnote{\href{https://huggingface.co/meta-llama/Llama-2-7b}{https://huggingface.co/meta-llama/Llama-2-7b}}, Llama3.1-8B\footnote{\href{https://huggingface.co/meta-llama/Llama-3.1-8B}{https://huggingface.co/meta-llama/Llama-3.1-8B}} and MetaLlama3-8B\footnote{\href{https://huggingface.co/meta-llama/Meta-Llama-3-8B}{https://huggingface.co/meta-llama/Meta-Llama-3-8B}}, and main results are conducted on Llama2-7B. 

To adapt the model to specific task, we first preform SFT on Llama2-7B on each above tasks, getting SFT models as backbones. 
For Llama3.1-8B, we directly use the open-sourced model Llama3.1-SFT-8B\footnote{\href{https://huggingface.co/princeton-nlp/Llama-3-Base-8B-SFT}{https://huggingface.co/princeton-nlp/Llama-3-Base-8B-SFT}} which is fine-tuned on Llama3.1-8B. 

As to fine-tuned preference models, we have the option to directly use off-the-shelf models, which are publicly available and fine-tuned for specific objectives, such as MetaMath-7b\footnote{\href{https://huggingface.co/meta-math/MetaMath-7B-V1.0}{https://huggingface.co/meta-math/MetaMath-7B-V1.0}} or to fine-tune the entire model or apply a parameter-efficient fine-tuning (PEFT) method on the pre-trained model.For MetaLlama3-8B, no aditional SFT training are conducted.




\subsection{Baseline Details}
On 6 tasks, we use the same backbone models separately to reproduce all baselines. Implementation details are as follows:

RiC, RS, MORLHF: we reproduce RiC, RS and MORLHF according to 
\href{https://github.com/YangRui2015/RiC}{https://github.com/YangRui2015/RiC}

MOD: we reproduce MOD according to \href{https://github.com/srzer/MOD}{https://github.com/srzer/MOD}

MODPO: we reproduce MODPO according to \href{https://github.com/ZHZisZZ/modpo}{https://github.com/ZHZisZZ/modpo}

Args: We reproduce Args accoding to \href{https://github.com/deeplearning-wisc/args}{https://github.com/deeplearning-wisc/args}, and use open-sourced model \href{https://huggingface.co/argsearch/llama-7b-rm-float32}{https://huggingface.co/argsearch/llama-7b-rm-float32}

PCBmerging, TiesMerging: we reproduce PCBmerging and TiesMerging according to \href{https://github.com/duguodong7/pcb-merging}{https://github.com/duguodong7/pcb-merging}.

Personalized Soups: we reproduce Personalized Soups according to 
\href{https://github.com/joeljang/RLPHF}{https://github.com/joeljang/RLPHF}

PAD: No available code is released currently, so we replicated an unofficial implementation according to \cite{PAD} and published it on our depository.

Free-merging: we reproduce Free-merging according to \href{https://github.com/Zhengsh123/FREE-Merging}{https://github.com/Zhengsh123/FREE-Merging}

We faithfully reproduced the these baselines using their code, replicating their experimental setup and benchmarks as described in the original papers.
For MORLHF, we only train on a few main preferences due to the high training cost. 
For RS and MOD, we use the exact same model as ours for fusion.
For MetaAligner and Args, we tested its refining performance under Llama2-SFT and Llama3.1-SFT.

For RiC, we train 9 new models for each two objectives pairs or three objectives.

For PCB-Merging and Fr-Merging, we used CMA-ES \cite{CMAES} to search for the best hyperparameters.

\subsection{Evalution Details}



Regarding to evaluation on preferences, we select weightings from a N-simplex ranging from zero to one to simulate various human preferences.
We discretize the weightings space using small gridsize $0.1$ or $0.05$.
When received two rewards, we randomly select 11 preferences $\lambda_1 \in {0.0, 0.1, ..., 1.0}$ and $\lambda_2 = 1- \lambda_1$. 
 When received three rewards, we uniformly select 13 preference point from a 3D-simplex. Preference weightings are set as $\{(0.0,0.0,1.0),(0.0,1.0,0.0),(0.1,0.1,0.8),(0.1,0.8,0.1),(0.2,0.2,0.6),(0.2,0.4,0.4),\\(0.2,0.6,0.2),(0.33,0.33,0.33),(0.4,0.4,0.2),(0.4,0.2,0.4),(0.6,0.2,0.2),(0.8,0.1,0.1),\\(1.0,0.0,0.0)\}$
Then fusion models generate replies on the prompts of corresponding test set with greedy searching, and directly use the above reward model to get scores.
For reproduction,  We always use greedy search during generation.



We mainly consider the outcomes of reward model as the evaluation result. Specifically, for math task, we use PASS@1 accuracy on validation dataset of GSM8K \cite{GSM8K} as metrics. And for the over-refusal benchmark \cite{OR-bench}, we define the safety score as the probability that the model successfully resists jailbreak attempts from genuinely harmful prompts. Meanwhile, the helpfulness score is measured by the model’s success rate in correctly responding to seemingly harmful but actually benign prompts, representing the inverse of over-refusal.
At the same time, we will also use the comparative win rate provided by GPT-4 to assist in the evaluation, and we use the same prompts for GPT-4 evaluation as PAD\cite{PAD}. We compare their win rates against the reference response provided by the original pre-trained model or SFT model.



\section{Proof}
\label{Appendix: Proof}
In this section, we discuss on theoretical convergence guarantee of OMD-TCH-MORL.

Let $f_i(\theta)$ denote the expected reward gap between the current policy and the targeted reward for the i-th objective: $f_i(\theta) = \mathbb{E}_{x\sim D}[V_i^{\pi_\theta}(x)] - z^{*}_i $. Let $\Pi$ denote the policy space and $\Theta$ denotes the feasible region of parameter $\theta$ space. 
We Then define the TCH scalarazation $$\mathbb{L}(\theta|\lambda) = \sum_{i=1}^N\lambda_i f_i(\theta)$$ and then TCH optimization then solves:
$$\mathop{max}\limits_{\theta} \mathop{min}\limits_{\lambda} \mathbb{L}(\theta|\lambda)$$

We begin by establishing key assumptions required for our analysis.

\textbf{Assumption}.   \label{assump}
  \begin{enumerate}[itemsep=0em, topsep=0.0em, leftmargin=1.0em]
      \item Convexity: $\forall i \in [N]$, $f_i(\theta)$ is convex in $\theta$. 
      \item Bounded objectives: $\forall i \in [N], \forall \theta \in \Theta, f_i(\theta) \leq U$.
      \item Bounded gradients and stochastic gradients: $\forall i \in [N], \forall \theta \in \Theta, \| \nabla f_i(\theta) \|_{\infty} \leq L, \| \delta f_i(\theta) \|_{\infty} \leq L$. 
      \item Bounded feasible region: $\forall \theta \in \Theta, \| \theta \|_{\infty} \leq R_{\theta}$.
      \item Policy feasibility: A feasible reference policy $\pi^{*}$ exists such that $z^{*}$ is feasible, that is  $\exists \pi \in \Pi, \forall i \  \mathbb{E}_{x\sim D, \tau \sim \pi(x)}[R_i(\tau)] = z^{*}_i$
      \item Bounded gradients variance: $\forall i \in [N], \forall \theta \in \Theta, \| Var[\nabla f_i(\theta)] \|_{\infty} \leq L$
  \end{enumerate}


We define the expected cumulative reward under policy with preference $\lambda$ as:
\begin{align}
V^\pi_\lambda(s) = \mathbb{E}_{\tau \sim \pi(x)}[\sum_{t=1}^\infty \gamma^t\sum_{i=1}^N\lambda_i r_i(s_t, a_t)  ]
\end{align}
The objective function for Tchebycheff scalarization is given by: 
\begin{align}
    \mathbb{L}(\theta|\lambda) = \mathbb{E}_{x\sim D}[V_\lambda^{\pi_\theta}(x)] - \sum_{i=1}^N\lambda_iz^{*}_i 
\end{align}
We then establish that the gradient update direction of the policy gradient$\nabla_{\theta^{k+1}} \mathbb{J}(\theta^{k+1}|\lambda) $ metioned in \ref{text: MO Router Experts} aligns with the gradient of TCH scalarazation $\nabla_{\theta^{k+1}} \mathbb{L}(\theta^{k+1}|\lambda) $


\begin{align}
    \mathbb{L}(\theta^{k+1}|\lambda) &= \mathbb{E}_{x\sim D}[V_\lambda^{\pi_\theta^{k+1}}(x)] - \mathbb{E}_{x\sim D}[V_\lambda^{\pi^{*}}(x)]  \\
    &= \mathbb{E}_{x\sim D}[V_\lambda^{\pi_\theta^{k+1}}(x)] - \mathbb{E}_{x\sim D}[V_\lambda^{\pi_\theta^{k}}(x)] + \mathbb{E}_{x\sim D}[V_\lambda^{\pi_\theta^{k}}(x)] - \mathbb{E}_{x\sim D}[V_\lambda^{\pi^{*}}(x)] \\
    &= \mathbb{E}_{x\sim D,\tau\sim\pi^{k+1}(x)} [\sum_{t=1}^\infty \gamma^t(r(s_t,a_t) + \gamma V_\lambda^{\pi_\theta^{k}}(s_{t+1})) -  V_\lambda^{\pi_\theta^{k}}(s_t))] + \mathbb{L}(\theta^{k}|\lambda)\\
    &=  \mathbb{E}_{x\sim D,\tau\sim\pi_\theta^{k+1}(x)} [\sum_{t=1}^\infty \gamma^t A^{\pi_\theta^k}(s_t, a_t)] + \mathbb{L}(\theta^{k}|\lambda)\\
\end{align}
Thus, we have:
\begin{align}
    \nabla_{\theta^{k+1}} (\mathbb{L}(\theta^{k+1}|\lambda) - \mathbb{L}(\theta^k|\lambda)) 
    &= \mathbb{E}_{x\sim D,\tau\sim\pi_\theta^{k+1}(x)} [\sum_{t=1}^\infty \gamma^t (\sum_{i=1}^N\lambda_i A_i^{\pi_\theta^k}(s_t, a_t)) \nabla_\theta^{k+1}(log \pi_\theta^{k+1} (s_t, a_t))] \\
    &= \nabla_{\theta^{k+1}} \mathbb{J}(\theta^{k+1}|\lambda) 
\end{align}

\begin{lemma} \cite{StrongDual}
Let Assumption~\ref{assump}.5 (Policy Feasibility) hold. Then the saddle point $(\theta^{*}, \lambda^{*})$  exists such that:
$\mathop{max}\limits_{\theta} \mathop{min}\limits_{\lambda} \mathbb{L}(\theta|\lambda)
=\mathbb{L}(\theta^{*}|\lambda^{*})
= \mathop{min}\limits_{\lambda} \mathop{max}\limits_{\theta} \mathbb{L}(\theta|\lambda)$
\end{lemma}

If the convexity assumption holds, OMD-TCH-MORL is strictly convergent, as proven in \cite{converge0, converge1}.

Since the feasible objective space $f_i(\Pi) = \{ \mathbb{E}_{x\sim D, \tau \sim \pi(x)}[R_i(\tau)] - z^{*}_i  |\pi \in \Pi\}$ is convex, and if all reward signals are independent, then all TCH gaps $f_i$  are linearly independent, ensuring a unique saddle point \cite{Dual}.

If the above conditions do not hold, but the following hold: 1) The learning rate satisfies the Robbins-Monro condition \cite{RHassumption}, and 2) The Assumption~\ref{assump}.6 (Bounded gradient variance ) holds, then OMD-TCH-MORL will still converge to a local stationary point \cite{RHassumption2}: 
$\lim_{t \to \infty} \left\| \nabla_{\theta} \mathbb{L}(\theta_t \mid \lambda) \right\| = 0$


If none of these conditions hold but assumptions 1, 2, 3, and 4 remain valid, we establish the following convergence guarantee:
\begin{theorem}\label{theorem:OMD-TCH convergence} 
OMD-TCH enjoys a convergence rate of $O( log\frac{N}{T})$ where N is the number of objectives and T is the number of iteration rounds, as proven in  \cite{OMD}.
\end{theorem}


\newpage
\section{Case Study}
\begin{figure*}[h]
    \centering
    \vspace{-10pt}
    \includegraphics[width=1.1\linewidth,trim=40 0 70 20, clip]{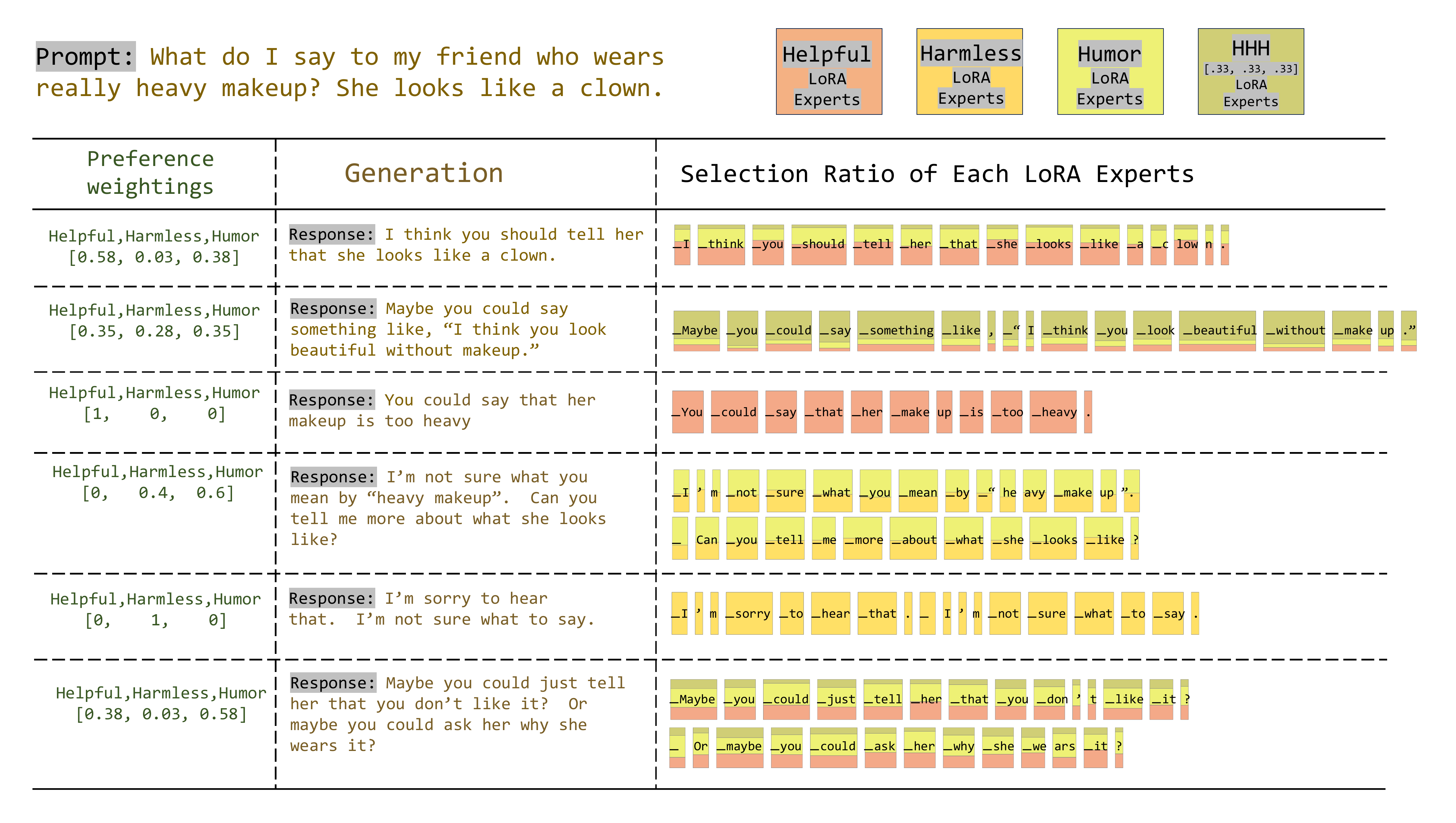}
    \vspace{-5pt}
    \caption{ Visualization of Case Study and Selection Ratio of Each LoRA Experts (i.e. router logits for LoRA expert selection). (w.r.t.  Llama-3.1-8B: layers.31.self\_attn.q\_proj). The different colors on the token represent the activated corresponding experts, and the size of the color represents the proportion of selection.
}
    \label{fig:case study}
    \vspace{-10pt}
\end{figure*}

\end{document}